\documentclass[10pt,twocolumn,letterpaper]{article}

\usepackage[pagenumbers]{cvpr} %

\usepackage{booktabs}
\usepackage{tabularx}      %
\usepackage{subcaption}    %

\definecolor{cvprblue}{rgb}{0.21,0.49,0.74}
\usepackage[pagebackref,breaklinks,colorlinks,allcolors=cvprblue]{hyperref}

\def\paperID{11759} %
\def\confName{CVPR}
\def\confYear{2026}

\title{Objects in Generated Videos Are Slower Than They Appear:\\ Models Suffer Sub-Earth Gravity and Don't Know Galileo's Principle...for now}
\author{Varun Varma Thozhiyoor$^1$\hspace{10pt} Shivam Tripathi$^1$\hspace{10pt} Venkatesh Babu Radhakrishnan$^1$\hspace{10pt} Anand Bhattad$^2$ \\\vspace{-8pt}\\
$^1$Indian Institute of Science, Bangalore\hspace{20pt}$^2$Johns Hopkins University\\\vspace{-8pt}\\
\href{https://gravity-eval.github.io}{Project: https://gravity-eval.github.io}}

\begin{document}

\twocolumn[{%
\renewcommand\twocolumn[1][]{#1}%
\maketitle\vspace{-2em}
    \includegraphics[width=1\textwidth]{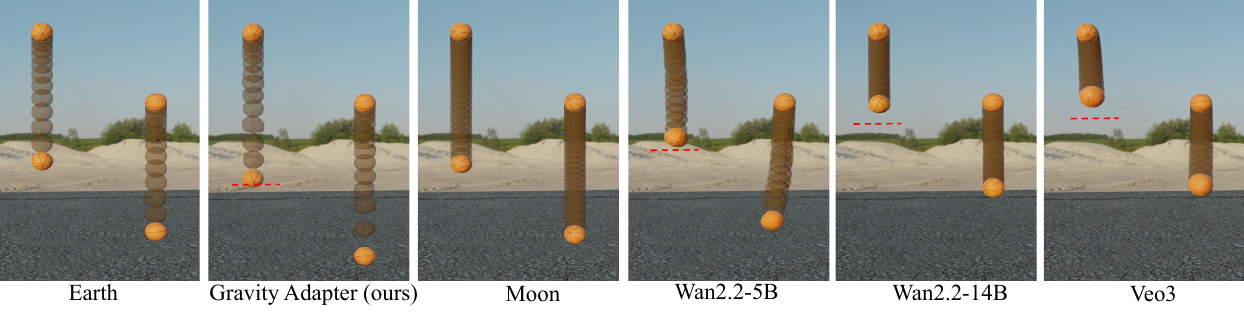}
    \vspace{-19pt}
    \captionsetup{type=figure}
    \vspace{-2pt}
    \captionof{figure}{\textbf{Video generators produce physically implausible slow-motion falls and most fail to understand that objects fall at equal rates.} We visualize two identical balls dropped simultaneously from different heights using stroboscopic (time-lapse) composites, tracking motion until the lower ball impacts the ground. \textbf{Two failures emerge:} (1)~\textbf{Galileo's principle violations:} Under Galileo's principle, both balls should fall equal distances in equal time regardless of starting height. Red dashed lines mark the expected position of the higher ball if both fell at equal rates. Earth (leftmost; simulation from Blender) shows the higher ball reaching this expected position, confirming correct physics. Moon reference ($g \approx 1.6\,\mathrm{m/s^2}$) shows both balls falling slower but preserving equal-rate progression. In contrast, Wan~14B and Veo3 show the higher ball severely lagging behind the red line. It traveled far less distance than the lower ball despite falling for the same time, violating the fundamental principle that gravitational acceleration is universal. Even Wan~5B shows noticeable lag. (2)~\textbf{Severe under-acceleration:} The spacing between successive ball positions indicates effective acceleration; wider spacing means higher acceleration. Most models exhibit compressed spacing comparable to Moon ($1.6\,\mathrm{m/s^2}$) or Mars ($3.7\,\mathrm{m/s^2}$) rather than Earth's $9.81\,\mathrm{m/s^2}$, revealing motion dramatically slower than terrestrial physics. Our Gravity Adapter (second panel), fine-tuned on Wan~5B with just 100 examples, corrects both failures by bringing the higher ball to the expected position and improving Earth-like spacing. 
    }
    \label{fig:teaser}
    \vspace{3mm}
}]

\maketitle
\begin{abstract}
Video generators are increasingly evaluated as potential world models, which requires them to encode and understand physical laws. We investigate their representation of a fundamental law: gravity. Out-of-the-box video generators consistently generate objects falling at an effectively slower acceleration. However, these physical tests are often confounded by ambiguous metric scale. We first investigate if observed physical errors are artifacts of these ambiguities (e.g., incorrect frame rate assumptions).  We find that even temporal rescaling cannot correct the high-variance gravity artifacts. To rigorously isolate the underlying physical representation from these confounds, we introduce a unit-free, two-object 
protocol that tests the timing ratio $t_1^2/t_2^2 = h_1/h_2$, a relationship independent of $g$, focal length, and scale. This relative test reveals violations of Galileo's equivalence principle. We then demonstrate that this physical gap can be partially mitigated with targeted specialization. A lightweight low-rank adaptor fine-tuned on only 100 single-ball clips raises $g_{\mathrm{eff}}$ from $1.81\,\mathrm{m/s^2}$ to $6.43\,\mathrm{m/s^2}$ (reaching $65\%$ of terrestrial gravity). This specialist adaptor also generalizes zero-shot to two-ball drops and inclined planes, 
offering initial evidence that specific physical laws can be corrected with minimal data.
\vspace{-8pt}
\end{abstract}  
    
\section{Introduction}
Two balls, a basketball and a tennis ball, are dropped from the same height, side by side. Assuming no air resistance, which hits the ground first? The answer is neither: they strike the ground simultaneously. Galileo demonstrated this principle over 400 years ago, and it remains one of the most fundamental predictions of Newtonian mechanics~\cite{newton1999, drake2003galileo}. In this paper, we investigate whether state-of-the-art video generators capture the fundamental physical law of gravity.

Galileo's principle is qualitative. A more rigorous, quantitative test is \emph{how} objects fall: the time to impact is proportional to the square root of the height ($t \propto \sqrt{h}$). But testing this in a generative model is not straightforward. Recent work shows that while models learn rich 3D representations~\cite{elbanani2024probing, bhattad2023stylegan, du2023generative, zhan2024general}, this internal geometry is inherently scale-ambiguous, and video frame rates provide no absolute time reference. Any rigorous test of physical law must therefore eliminate these confounds.

Prior benchmarks audit physics through trajectory fitting~\cite{li2025pisa}, conservation laws~\cite{zhang2025morpheusbenchmarkingphysicalreasoning}, or broad qualitative suites~\cite{bansal2024videophyevaluatingphysicalcommonsense, motamed2025generative, kang2025how, wiedemer2025video}. While valuable, these approaches either require camera calibration, rely on approximate heuristics, or optimize for absolute positional accuracy in a calibrated space -- conflating the law of gravity with its parameterization. None directly measures whether a model's representation of gravity quantitatively matches Earth's $g=9.81\,\mathrm{m/s^2}$ in a way that is robust to the scale and time-base ambiguities inherent in generated video.

We introduce a unit-agnostic protocol based on relative timing between simultaneously falling objects. Because generated videos provide \emph{no} guarantee that the model's internal `meter' aligns with physical units or that frame intervals correspond to absolute time, we must eliminate reliance on explicit units. Under a static pinhole camera with zero initial velocity, the ratio of impact times satisfies
\vspace{-5pt}
\[ \frac{t_1^2}{t_2^2}=\frac{h_1}{h_2}, \\\vspace{-2pt}\]
where metric scale, focal length, time base, and gravitational constant all cancel---yielding a calibration-free diagnostic of physical understanding.

We find that, out of the box, most models violate Galileo's principle that objects fall at the same rate; see Figure~\ref{fig:teaser}. In our single-ball drop experiments ($h \in [0.5,4.0]\,\mathrm{m}$), impacts are consistently late across all tested video generators (Wan~5B, Wan~14B, Veo3, Cosmos~2B, Cosmos~14B). Interpreted in their native time gauge, the implied effective gravity $g_{\mathrm{eff}}$ ranges from $0.38$ to $2.27\,\mathrm{m/s^2}$, far below Earth's $9.81\,\mathrm{m/s^2}$; balls therefore fall much more slowly than real-world gravity would predict, exhibiting an effective ``sub-Earth'' gravity. A straightforward temporal rescaling can move the \emph{mean} $g_{\mathrm{eff}}$ closer to $9.81\,\mathrm{m/s^2}$, but the resulting distributions remain broad with large variance and heavy tails, indicating substantial inconsistency across scenes. Moreover, our two-object protocol reveals systematic, nonzero time differences $\Delta t$ between the two balls when they traverse the same vertical distance. This relative-timing discrepancy is invariant to any global time rescaling and thus indicates a genuine violation of Galileo's principle. Even more surprisingly, larger models (Wan~14B, Cosmos~14B) exhibit \emph{slower} motion than their smaller counterparts (Wan~5B, Cosmos~2B), contradicting the assumption that scale alone improves physical consistency.

To repair this error, we train a lightweight LoRA adapter on just 100 single-ball sequences. Applied to Wan~5B, it increases $g_{\mathrm{eff}}$ on average from $1.81\mathrm{m/s^2}$ to $6.43\mathrm{m/s^2}$---reaching $65\%$ of terrestrial gravity on average. Interestingly, this specialist generalizes zero-shot to two-ball drops and inclined planes despite never seeing these scenarios during LoRA training. In summary, objects in generated videos are slower than they appear. Video generators excel as generalists but struggle as physics engines; our adapter supplies targeted specialization with minimal data. Just as Galileo showed that the falling rate is independent of mass, we show that physical accuracy is independent of model scale; it requires targeted correction, not raw capacity.

\noindent\textbf{Contributions.}
\textbf{(1)}~We introduce a unit-free, two-object measurement protocol that isolates gravitational acceleration while being invariant to camera scale and frame rate, providing a rigorous diagnostic for physical consistency in video generators.
\textbf{(2)}~We quantify systematic physics violations across state-of-the-art models, revealing effective gravity at $5\text{--}20\%$ of Earth's value in their native time gauge and showing that larger models can be \emph{less} physically accurate than smaller ones.
\textbf{(3)}~We show that a 100-example LoRA adapter on Wan~5B and Wan~14B partially improves these gravity metrics and shows generalization to unseen scenarios, demonstrating the feasibility of correcting specific physical laws with minimal data.

\section{Related Works}
\textbf{Physics Evaluation in Video Generators.} 
Recent benchmarks evaluate physics understanding through pixel-to-ground-truth matching~\cite{li2025pisa, motamed2025generative}, human/VLM-as-a-judge~\cite{bansal2024videophyevaluatingphysicalcommonsense, guo2025t2vphysbenchfirstprinciplesbenchmarkphysical, li2025worldmodelbenchjudgingvideogeneration, gu2025phyworldbenchcomprehensiveevaluationphysical}, or trajectory modeling~\cite{zhang2025morpheusbenchmarkingphysicalreasoning, kang2025how}. However, these approaches face a fundamental challenge: scale and time ambiguity. Monocular vision~\cite{eigen2014depth, zhou2017unsupervised} and neural rendering~\cite{ahn2024alpha} cannot recover absolute metric scale without calibration. Video generators inherit this limitation: their internal representations are unitless, and frame rates provide no absolute time base~\cite{ho2022video}. Yet existing physics benchmarks implicitly assume calibrated measurements, conflating the physical law itself with its parameterization (e.g., testing $g = 9.81\,\mathrm{m/s^2}$ specifically rather than the square-root scaling law). While VLMs enable broad qualitative scope~\cite{garrido2025intuitivephysicsunderstandingemerges, motamed2025travlrecipemakingvideolanguage}, they suffer from hallucinations and prompt shortcuts~\cite{motamed2025travlrecipemakingvideolanguage, sanli2025modelseparateyolkswater}, making them unreliable for quantitative physical laws. We address this by introducing unit-free relative measurements that isolate physical principles from metric parameterization.

\textbf{Falling Objects.} 
For free-fall specifically, PISA~\cite{li2025pisa} optimizes trajectory IoU using 5,000 simulated videos with reward-based fine-tuning, while Morpheus~\cite{zhang2025morpheusbenchmarkingphysicalreasoning} compares extracted trajectories against physics simulations. However, neither quantifies whether models understand gravity's fundamental properties: the square-root time-height relationship ($t \propto \sqrt{h}$) or Galileo's equivalence principle (all objects fall at equal rates). They only measure deviation from ground-truth videos with implicit, unspecified gravity values. Kang et al.~\cite{kang2025how} train 2D models from scratch to test generalizability but similarly do not isolate the physical law from scale ambiguity. Our two-ball protocol eliminates these confounds by testing timing ratios ($t_1^2/t_2^2 = h_1/h_2$), where scale, focal length, and gravity all cancel. This provides a calibration-free test of gravity.

\textbf{Probing Knowledge in Generative Models.} 
Recent work investigates what visual knowledge generative models encode, including depth, normals, semantics, and object relationships~\cite{elbanani2024probing, zhan2024general, bhattad2025visual}, as well as disentangled representations of lighting and geometry~\cite{bhattad2023stylegan, du2023generative, bhattad2024stylitgan, xing2025luminet}. Sarkar et al.~\cite{sarkar2024shadows} revealed that diffusion models systematically fail at projective geometry, mispredicting shadows and vanishing points despite encoding 3D cues. Our work extends this inquiry from static geometry to temporal dynamics, uncovering systematic failures in gravitational physics. Critically, we find that a 100-example LoRA corrects these failures and generalizes zero-shot to unseen scenarios, suggesting physical laws occupy sparse, learnable subspaces within foundation models. This connects interpretability research with physics-guided generation, showing that models encode latent physical knowledge that requires targeted activation rather than learning from scratch.

\textbf{Physics-Guided Generation.} 
Methods for improving physical correctness follow two paradigms: distilling physics through post-training alignment~\cite{li2025pisa, zhang2025videorepa} or explicitly simulating scenes with physics engines~\cite{liu2024physgen, Yuan_2025_NewtonGen, wang2025physctrl, chen2025physgen3d}. Post-training approaches like PISA require massive compute, while simulation-based methods depend on accurate scene estimation and cannot handle complex real-world scenarios. Both align with the intuitive physics hypothesis~\cite{battaglia2013simulation} that physical understanding emerges from observing dynamics. However, our findings reveal that despite exposure to vast video data, models learn correlational patterns without internalizing fundamental rules. They systematically violate both acceleration laws and Galileo's equivalence principle. Unlike prior work, our lightweight LoRA adapter achieves substantial correction and generalizes zero-shot to two-ball drops and inclined planes, demonstrating that targeted specialization can efficiently bridge the gap between visual plausibility and physical correctness.

\textbf{Intuitive Physics and World Models.} 
Understanding physical dynamics from vision is central to AI systems that model the world. The intuitive physics hypothesis~\cite{battaglia2013simulation} suggests observers learn physical constraints through internal simulation. Benchmarks like IntPhys~\cite{riochet2018intphys} and Physion~\cite{bear2021physion} evaluate physical prediction in vision systems, while world models~\cite{craik1967nature, ha2018world, hafner2023mastering, micheli2023transformers, bruce2024genie} learn environment dynamics for planning in reinforcement learning. However, prior work focuses on discriminative prediction or task-specific environments rather than general-purpose video synthesis. Our work reveals that while video generators excel at visual plausibility, they systematically fail at basic physical laws. A system believing objects fall at Moon gravity cannot serve as a reliable world model. Our findings suggest current video generators learn correlational patterns without internalizing physical constraints, though our adapter demonstrates targeted specialization can partially bridge this gap.

\section{Experimental Setup}
\label{sec:experimental_setup}

We investigate how well modern video generators capture gravitational physics in the canonical setting of falling objects. Our evaluation requires controlled generation and measurement; we describe our synthetic benchmark, the models tested, and our tracking-based evaluation protocol.

\subsection{Synthetic Benchmark}

\paragraph{Rendering.}Each video in our benchmark is created synthetically using Blender. We use various standard sports balls as falling objects, which should be well-represented in the models' training data. Balls are simulated to be dropped from random heights in the range $[0.5, 4.0]\,\mathrm{m}$, and the initial height is recorded. Note that we are only interested in evaluating the gravity of falling objects. Therefore, both the ground-truth data used for evaluation and the data used to train the gravity adapter omit any physical simulation of bounces. Since bouncing behavior depends on material properties of both the object and the ground surface, we choose not to simulate it. 

The camera is fixed at a sufficient distance and height to capture the entire trajectory of the ball and is positioned perpendicular to the falling motion. The scene is lit, and the background is occupied by HDRI maps. We use indoor and outdoor HDRI maps with a variety of different ground materials to match the setting. Backgrounds are chosen to match the scale of the ball---objects in the background do not appear too large or too small compared to the ball. We include samples with backgrounds having objects, to provide more context, as well as samples with more plain backgrounds that provide less context for more challenging generations(additional objects provide scale context). Videos are rendered at $1280{\times}720$ resolution for 2 seconds at 24~fps, yielding 48 frames per sequence.

We generate $N$ videos with heights uniformly sampled from $[0.5, 4.0]\,\mathrm{m}$. 
The dataset includes $K$ different ball types (basketball, soccer ball, tennis ball, volleyball, baseball) 
across $M$ unique HDRI environments. For the adapter training (Sec.~\ref{sec:adapter}), 
we use 100 single-ball sequences; all other experiments use held-out test data.

The camera uses a focal length of $f = 50,\mathrm{mm}$ and is positioned between 1.0 and 8.0 meters from the drop zone. Its height is set to half the drop height, with an additional offset sampled from the range $[-0.5,,0.5]$ meters. This configuration ensures that perspective distortion remains minimal across the falling trajectory.

\vspace{-10pt}
\paragraph{Single-Ball Protocol.}  We first quantify the effective gravitational acceleration that models implicitly use by measuring fall time vs. height. This establishes a baseline before we eliminate scale ambiguity with our unit-free two-ball protocol. We generate 75 sequences with drop heights uniformly sampled from $[0.5, 4.0]\,\mathrm{m}$. This tests whether models produce trajectories consistent with $t = \sqrt{2h/g}$, allowing us to measure effective gravity $g_{\mathrm{eff}} = 2h/t^2$.

\begin{figure*}[t]
\centering

\begin{subfigure}{0.4\textwidth}
    \includegraphics[width=\linewidth]{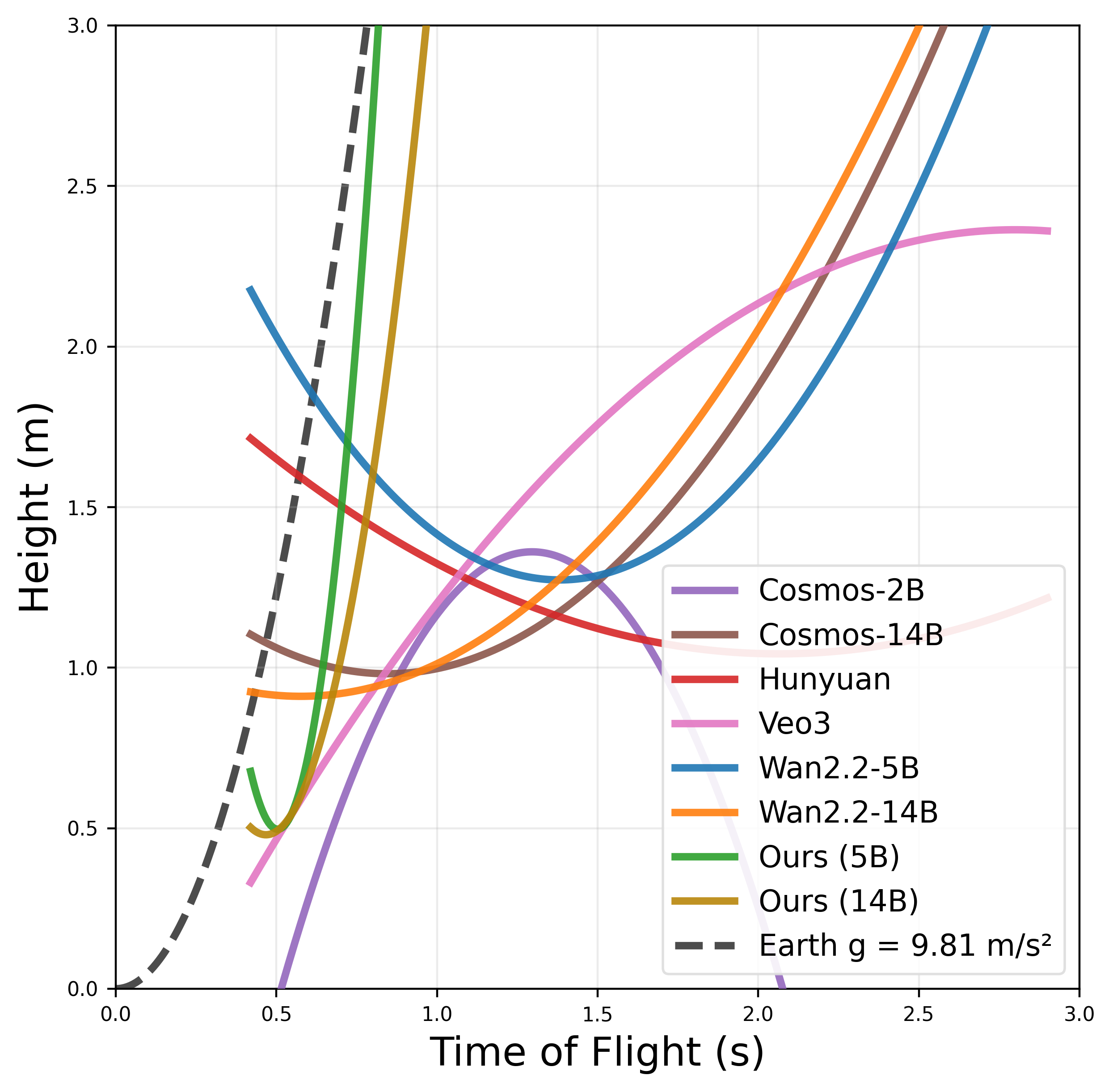}
    \caption{Original}
    \label{fig:og_single_ball_plot}
\end{subfigure}
\hspace{2em}
\begin{subfigure}{0.4\textwidth}
    \includegraphics[width=\linewidth]{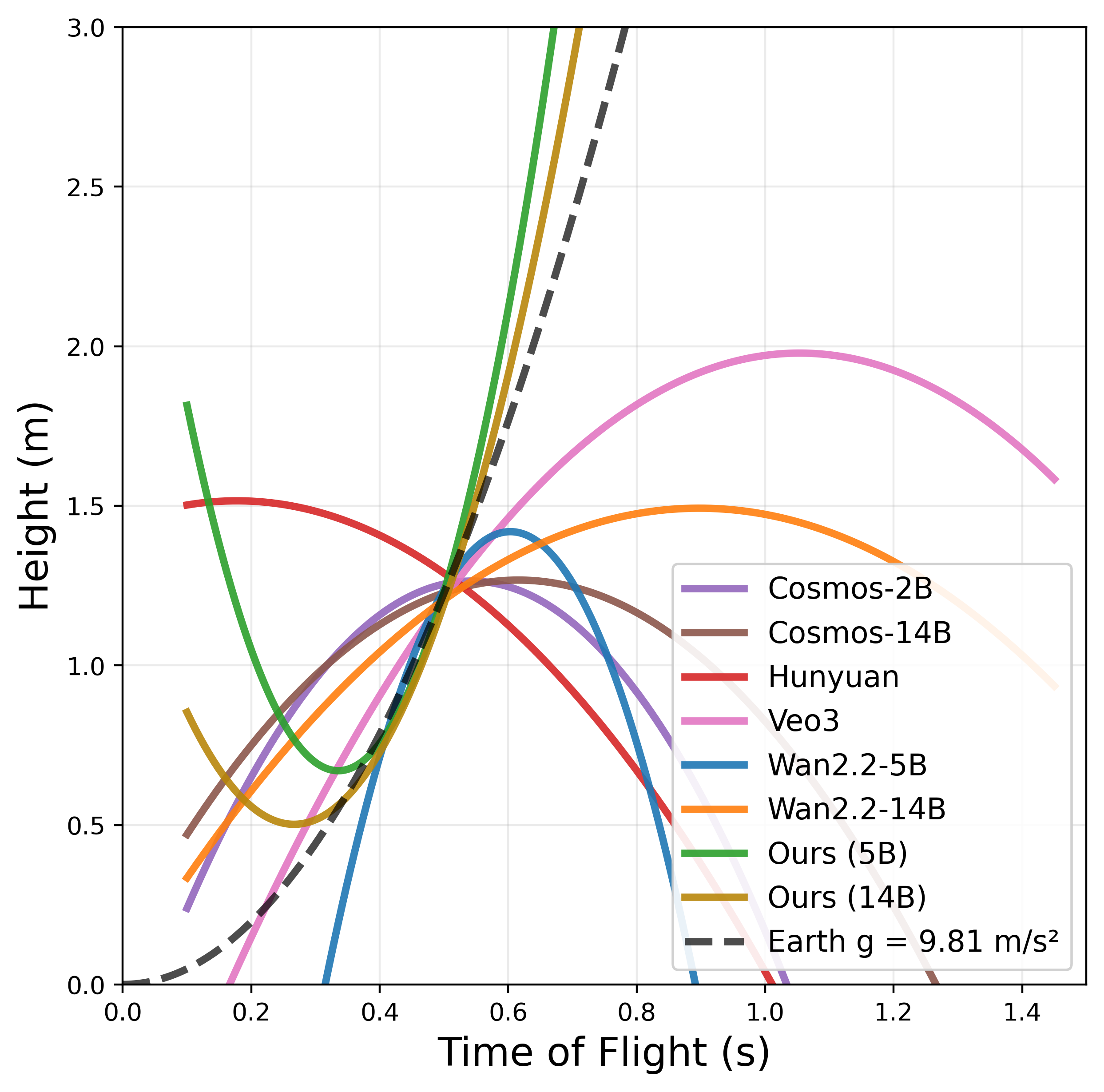}
    \caption{Mean Scaling}
    \label{fig:time_scaled_single_ball_plot}
\end{subfigure}

\caption{
\textbf{Effect of time-scaling on $h$–$t$ relationships.}
\textbf{(a)} We plot $h$ versus $t$ for all models. We repeat each test example with 4 seeds and fit polynomials through the means. The gray dashed line indicates terrestrial motion. All models systematically under-accelerate, and none obey the square root scaling law of time with height. The Gravity Adapters (green, gold) substantially improves Wan 5B and Wan 14B towards correct gravity.
\textbf{(b) Mean time scaling.} We compute a Mean time scalar using a subset of random 30 samples from our dataset, which scales the effective time of the 30 samples to better match the ground truth time. The Mean time scalar, when applied to the second subset of 45 samples, brings the mean effective gravity closer to 9.81, m/s2 for many models, but the variance remains high indicating that under-acceleration is not simply a frame rate artifact.
}
\label{fig:single_ball_plot_grid}
\end{figure*}

\vspace{-10pt}
\paragraph{Two-Ball Protocol.}
To eliminate metric scale and time-base ambiguity, we generate 50 sequences with two (identical) balls dropped simultaneously from different heights $h_1, h_2$ within the same frame. To ensure both balls experience identical perspective effects, they are positioned 
at the same distance from the camera and separated horizontally by ball-diameter with an offset of 0.5 meters to prevent occlusion. This guarantees that any scale ambiguity affects both balls equally, 
making the timing ratio truly unit-free.
Height ratios $h_1/h_2$ range from 0.25 to 3.5. The relative timing should satisfy $t_1^2/t_2^2 = h_1/h_2$, which cancels gravity, scale, focal length, and frame rate---providing a unit-free test of physical understanding.

\subsection{Models Evaluated}

We evaluate models with Image-to-Video capability to constrain the generations to fall from a known height specified using a conditioning image. Experiments are conducted on Wan~5B and Wan~14B~\cite{wan2025}, Veo3~\cite{Google_Veo3_2025}, Hunyuan~\cite{kong2024hunyuanvideo}, and Cosmos~2B and Cosmos~14B~\cite{nvidia_cosmos_predict2}. For each model, we supply the first frame of the video and a text prompt as initial conditions. We generate videos at resolutions and frame rates that match the ground-truth as closely as possible. For models supporting multiple settings, we select the configuration nearest to the target specification of $1280{\times}720$ at 24~fps. For cases where the model does not support 2-second generations, like Veo3, we choose a generation time close to 2 seconds. We use consistent text prompts across models (see \ref{sec:prompt_structure})
to isolate physics understanding from prompt sensitivity.

\subsection{Measurement Protocol}
Generated videos are resized and adjusted to $1280{\times}720$, 24~fps. We track the ball using SAM2~\cite{ravi2024sam2}, initialized with Blender centroids. Impact time $t$ is determined when: (1) the bottom of the ball drops below a point located one ball-radius (in pixels) above the ground threshold $y_{\mathrm{ground}}$, and (2) vertical velocity drops below $\varepsilon = 1\,\mathrm{pixels/frame}$. The ground threshold prevents false detections for hovering balls; the velocity threshold accounts for impact deformation. For single-ball drops, we compute $g_{\mathrm{eff}} = 2h/t^2$ and report mean, median and range. For two-ball drops, we compute timing ratios $t_1^2/t_2^2$ and compare against the theoretical $h_1/h_2$. We manually verify extreme acceleration results above $50\mathrm{m/s^2}$ and exclude any samples that are found to be incorrectly evaluated because of SAM failures.

\begin{figure*}[t!]
  \centering
\includegraphics[width=0.95\textwidth,height=0.45\textheight,keepaspectratio]{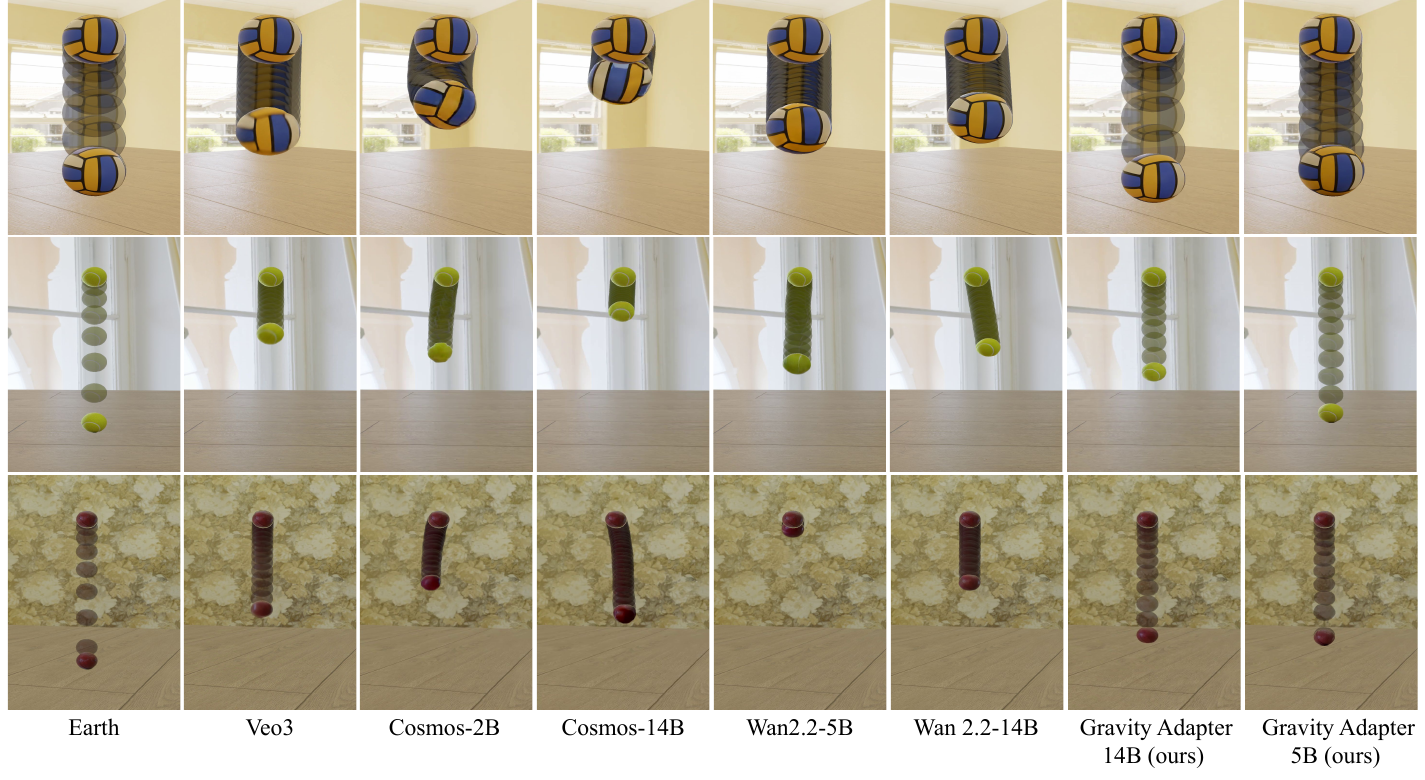}
\\\vspace{-10pt}
    \caption{\textbf{Scaled Single-ball drops reveal systematic under-acceleration across all models.} Stroboscopic composites (left) visualize ball positions at equal time intervals from release. The panels show the trajectories performed by each model during the time it takes a ball falling under $9.8 {m/s^2}$ to reach the ground, scaled by MTS (Tab.~\ref{tab:time_scaled_single_ball_table}). All models showcase severe under-acceleration (easily visible in the compressed spacing in the composites). The  Gravity Adapters (seventh and eighth column) substantially improves Wan 5B and Wan14B toward terrestrial dynamics. 
    }
\label{fig:single_ball_quali}
\vspace{-10pt}
\end{figure*}

\section{Results}
\label{sec:results}

\begin{table*}[t!]
\caption{\textbf{Effect of time-scaling on $g_{\mathrm{eff}}$ estimation across models.}
\textbf{(a) Original.} We report effective gravity values computed as 
$g_{\mathrm{eff}} = 2h/t^2$ (m/s$^2$). The ground truth is $9.81$ m/s$^2$. 
All models under-accelerate, and Gravity Adapters consistently reduce this deficit. 
Reported mean values are averaged over four random seeds and all test examples. Median and Range values are across all seeds and test samples.
\textbf{(b) Mean time scaling.} A global scalar (MTS) is estimated from a 
30-sample subset and applied to a disjoint 45-sample split. This shifts several models 
closer to $9.81$ m/s$^2$, but variance remains high. 
}
\centering
\begin{subtable}{0.98\columnwidth}
\centering
\caption{Original}
\label{tab:og_single_ball_table}
\tiny
\begin{tabularx}{0.95\linewidth}{lccc}
\toprule
\textbf{Model} & \textbf{Mean ($\mathrm{m/s^2}$)} & \textbf{Median ($\mathrm{m/s^2}$)}  & \textbf{Range ($\mathrm{m/s^2}$)}  \\
\midrule
 Cosmos~2B~\cite{nvidia_cosmos_predict2}&  \textbf{1.85} & 1.30 &  [0.23, 14.18]  \\
Cosmos~14B~\cite{nvidia_cosmos_predict2}&  1.51&  1.01&  [0.24,10.31]   \\
 Hunyuan~\cite{kong2024hunyuanvideo}&  1.97&  1.15&   [0.23,15.84] \\
 Veo3~\cite{Google_Veo3_2025}&  2.27 &  2.08& [0.28, 6.66] \\
 Wan~5B~\cite{wan2025}&  1.81&  \textbf{1.24} &   [0.26,8.26] \\
 Wan~14B~\cite{wan2025}&  2.18& 1.19 & [0.27, 59.98]     \\
 \midrule
Gravity Adapter~5B~\cite{hu2022lora}&  \textbf{6.43} & 6.38 & [1.24, 16.64] \\
Gravity Adapter~14B~\cite{hu2022lora} &  \textbf{5.51} &  5.63&  [1.52, 11.67]  \\
\bottomrule
\end{tabularx}
\end{subtable}
\hfill
\begin{subtable}{0.98\columnwidth}
\centering
\caption{Mean-Time Scaled}
\label{tab:time_scaled_single_ball_table}
\tiny
\begin{tabularx}{0.95\linewidth}{lcccc}
\toprule
\textbf{Model} & \textbf{MTS} & \textbf{Mean}($\mathrm{m/s^2}$) & \textbf{Median} ($\mathrm{m/s^2}$) & \textbf{Range} ($\mathrm{m/s^2}$)\\
\midrule
  Cosmos~2B~\cite{nvidia_cosmos_predict2}&  2.43& 10.34 &  7.24 & [1.40, 69.96]\\
 Cosmos~14B~\cite{nvidia_cosmos_predict2}&  2.77 & 10.86 & 7.19 & [1.81, 76.04]\\
  Hunyuan~\cite{kong2024hunyuanvideo}&  2.63 & 13.85 & 7.06 & [1.55, 104.64] \\
  Veo3~\cite{Google_Veo3_2025}& 2.12 & 9.39 & 8.5 & [1.26, 29.11] \\
  Wan~5B~\cite{wan2025}& 2.43 & 10.24 & 7.22 & [1.76, 49.15] \\
  Wan~14B~\cite{wan2025}& 2.56 & 13.78 & 7.78 & [1.75, 321.30] \\
  \midrule
Gravity Adapter~5B~\cite{hu2022lora}& 1.28 & 10.27 & 9.74 & [2.4, 26.67]\\
Gravity Adapter~14B~\cite{hu2022lora}& 1.36 & 10.00 & 10.10 & [2.82, 19.08]\\
\bottomrule
\end{tabularx}
\end{subtable}
\label{fig:single_ball_table_grid}
\end{table*}

\subsection{Single-Ball Drops: Sub-Earth Gravity}
\label{sec:single_ball}
Figure~\ref{fig:og_single_ball_plot} plots $h$ versus $t$ for all the models. Under correct physics, points should lie on the curve $t^2 = 2h/g$. All the models deviate from this reference.

Table~\ref{tab:og_single_ball_table} summarizes effective gravity measurements. No model reproduces Earth-like acceleration. 
Smaller models show less extreme under-acceleration (closer to Earth gravity) than larger models, though all remain far below $g = 9.81\,\mathrm{m/s^2}$. Paradoxically, scaling increases capacity but does not improve physical accuracy.
In the case of Wan~14B, even though it reports a higher mean value, the median and the corresponding distribution(\ref{fig:og_single_ball_histogram}) reveal that the majority of generations remain substantially under-accelerated. The increased mean is primarily driven by a small number of extreme samples.
Our lightweight Gravity Adapter, fine-tuned on just 100 sequences (Sec.~\ref{sec:adapter}), improves Wan~5B from $g_{\mathrm{eff}}{=}1.81$ to $g_{\mathrm{eff}}{=}6.43$ (\text{range:} 1.24 –16.64$\mathrm{m/s^2}$), demonstrating that targeted specialization can partially correct gravity.

\noindent\textbf{Does prompting help? No.}
To reduce any scale ambiguity for the models, we experiment with more detailed text prompts containing explicit height, diameter of the ball, distance of the camera from the ball, and height of the camera above the ground (see \ref{sec:prompt_structure} for prompt structure). However, we see no significant improvement in any of them (Tab \ref{tab:expanded_prompt}).

\noindent\textbf{Is under-acceleration a frame rate artifact? No.} We first test whether temporal stretching explains the low gravity values. Using the single-ball dataset, we split 75 samples into two subsets (30 and 45 videos). From the first subset, we compute each model's average time-scaling factor (\textit{MTS}) as $\text{mean}(t_{eff}/t_{gt})$. We then apply this global scaling factor to videos in the second subset ($1/MTS \times t_{eff}$), computing scaled gravity values $g_{scaled}$. Tab.~\ref{tab:time_scaled_single_ball_table} shows the results across samples and seeds. Since the absolute scale is unknown, we treat the single-ball experiment as a test of temporal consistency. If the model simply had a slow clock, a linear time scalar should fix the error. As Tab.~\ref{tab:time_scaled_single_ball_table} shows, it does not. This failure necessitates a unit-free metric to understand the root cause. 
See Fig.~\ref{fig:single_ball_quali} for qualitative examples demonstrating under-acceleration. We also tested other scaling baselines (see supplement): (1) per-scene mean-time scaling computed across multiple seeds, and (2) height-adjusted scaling for non-vertical trajectories. Neither approach yields substantial improvement, confirming that physics errors persist even with privileged per-sample correction.

\begin{table}[t]
\centering
\caption{\textbf{Use of expanded prompts.} Detailed prompts describing the scene with explicit parameters do not significantly affect the model's understanding of gravity. Wan~5B  increase marginally in $g_{eff}$. Veo~3 and Cosmos~2B show a decline.}
\label{tab:expanded_prompt}
\small
\resizebox{0.98\columnwidth}{!}{
\begin{tabular}{l|ccc|ccc}
\toprule
\textbf{Model} &
\multicolumn{3}{c}{\textbf{Base Prompt}} &
\multicolumn{3}{c}{\textbf{Expanded Prompt}} \\
\cmidrule(lr){2-4} \cmidrule(lr){5-7}
& \textbf{Mean} ($\mathrm{m/s^2}$) & 
\textbf{Median} ($\mathrm{m/s^2}$) &
\textbf{Range} ($\mathrm{m/s^2}$) & \textbf{Mean} ($\mathrm{m/s^2}$) & 
\textbf{Median} ($\mathrm{m/s^2}$) &
\textbf{Range} ($\mathrm{m/s^2}$) \\
\midrule

Wan~5B & 1.81 & 1.24 & [0.26, 8.26] & 2.06 & 1.37& [0.29, 6.80] \\
Veo3 & 2.27 & 2.08 & [0.28, 6.66] & 1.63  & 1.24 & [0.34, 4.94] \\
Cosmos~2B & 1.85 & 1.30 & [0.23, 14.18] & 1.25&0.83 & [0.27, 3.84] \\

\bottomrule
\end{tabular}
}
\vspace{-10pt}
\end{table}

\subsection{Two-Ball Drops: Failure of Galileo's Principle}
\label{double_ball}

Prior work evaluating physics in video generators\cite{li2025pisa, zhang2025morpheusbenchmarkingphysicalreasoning} predominantly tests single-object scenarios, comparing generated trajectories against ground truth. While such approaches can measure whether models follow text prompts accurately, they cannot determine whether models understand the underlying physical principles. A model might correctly match a single falling trajectory through a pattern 
recognition without understanding gravity.

We pose a more fundamental question: Do models understand Galileo's principle of gravitational equivalence -- that all objects fall at the same rate regardless of mass or starting height? To test this, we generate scenes with two identical balls dropped simultaneously from different heights $h_1 < h_2$. We use identical balls to eliminate depth-perception confounds that could arise from size differences. Under correct physics, both balls experience identical acceleration and thus fall equal distances in equal time. At the moment when the lower ball impacts at time $t_1 = \sqrt{\frac{2h_1}{g}}$, the higher ball should have fallen the same distance $d = h_1$, placing it at height $h_2 - h_1$. \textbf{This is a zero-ambiguity test}: either both balls fall together (pass) or they don't (fail).

\begin{figure*}[t]
    \centering
    \includegraphics[width=0.95\linewidth]{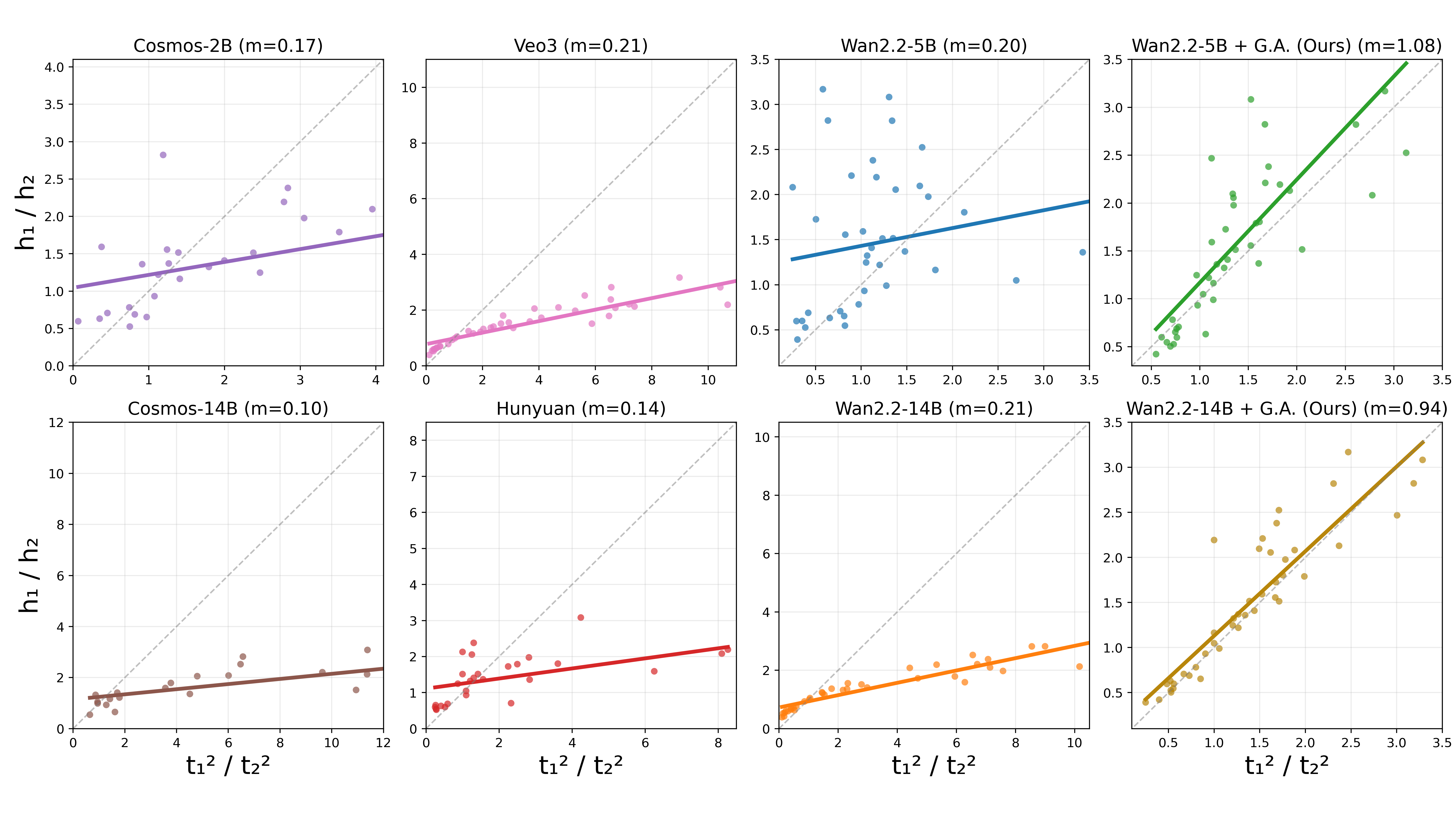}
    \\\vspace{-10pt}
    \caption{\textbf{Two-ball relative timing results.} We plot measured timing ratios $t_1^2/t_2^2$ against theoretical predictions $h_1/h_2$ across multiple height ratios. The gray dashed line indicates perfect agreement. All models deviate systematically, confirming that under-acceleration is not an artifact of scale estimation but reflects genuine physics and Galileo's principle violations. We also measure the slope (m) for each of them to understand the deviation.}
    \label{fig:double_ball_plot}
    \vspace{-10pt}
\end{figure*}

\begin{figure*}[t!]
  \centering
\includegraphics[width=0.95\textwidth,height=0.45\textheight,keepaspectratio]{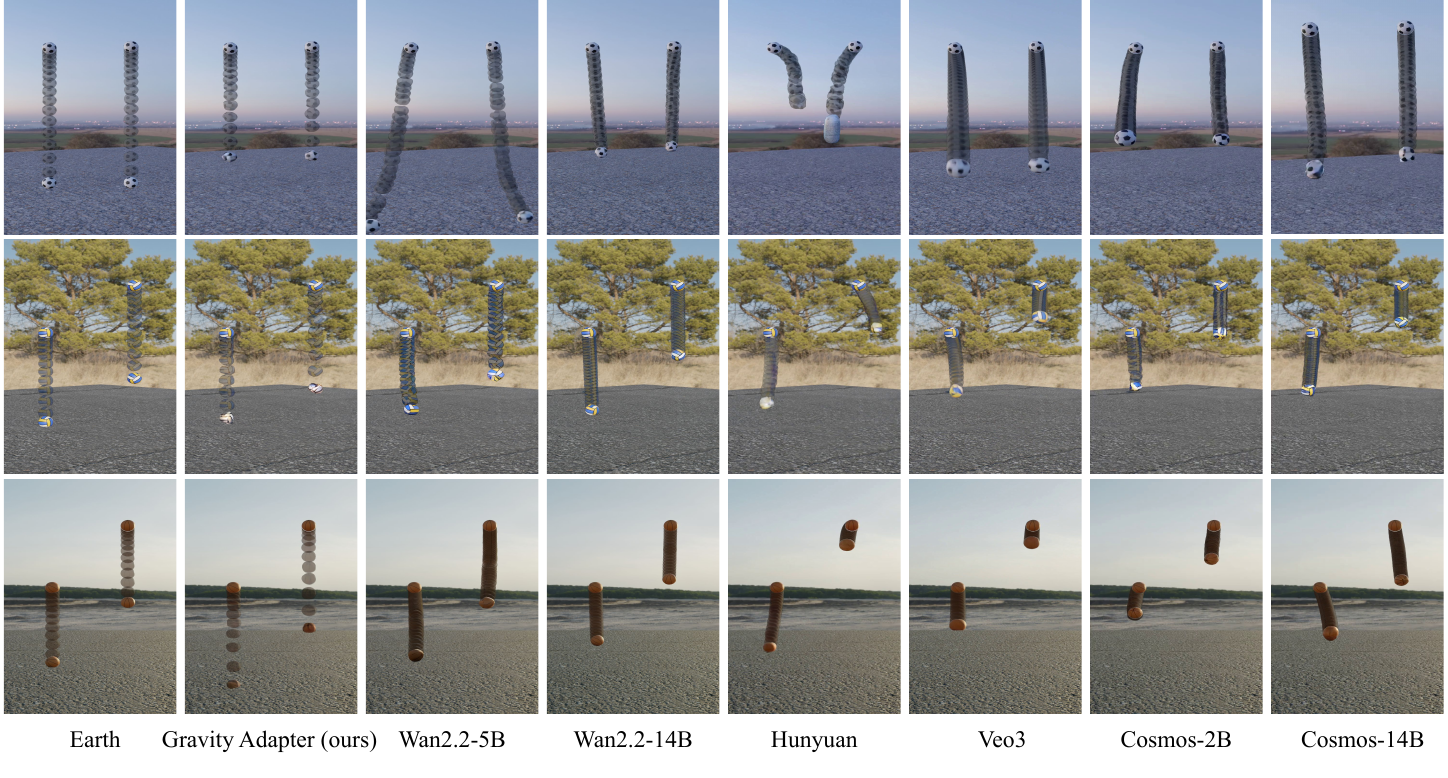}
\\\vspace{-10pt}
    \caption{\textbf{Most models fail Galileo's principle of gravitational equivalence.}  We freeze at the moment the lower ball (from height h$_1$) impacts the ground.  Under correct physics, both balls should have fallen equal distances in equal 
time, demonstrating that gravitational acceleration is universal (leftmost 
column: Ground Truth Earth). \textbf{Catastrophic failures:} Veo3 (sixth column) shows the higher ball barely moving while the lower ball lands—a complete violation of 400-year-old physics. Wan 14B, Veo3, and Cosmos 14B show similar failures with the higher ball remaining significantly elevated, suggesting these models believe in gravity depends on the starting height or object ordering. 
\textbf{Correction:} The Gravity Adapter (second column) finetuned on Wan 5B restores near-terrestrial acceleration \textit{and} perfects equal-rate falling, demonstrating that this fundamental physics deficit can be corrected with only 100 training examples.}
    \label{fig:double_ball_quali}
    \vspace{-10pt}
\end{figure*}

Figure~\ref{fig:double_ball_quali} reveals that \textbf{most models catastrophically fail}. Veo3, Wan 14B, and Cosmos 14B show the higher ball traveling an unequal distance compared to the lower ball, when the lower ball lands. This observation is confirmed by plotting $t_1^2/t_2^2$ versus $h_1/h_2$ across multiple height ratios as shown in Figure \ref{fig:double_ball_plot}. This is not just a uniform timing error; it reflects a failure to respect that gravitational acceleration is universal. Our two-ball experiments show that the upper and lower balls often take different times to traverse the same distance: in some models, the higher ball is effectively slower, in others faster, but in all cases the two objects experience different accelerations within the same scene, contradicting four centuries of physics.

To quantify deviations from Galileo's principle we measure the temporal lag between the two falling balls in units of video frames.
First, we record the time at which the lower ball reaches the ground, denoted as $t_1$, having covered a vertical pixel distance $y_1$. 
We then measure the time $t_2$ taken by the upper ball to traverse the same distance, and compute the deviation $\Delta t = t_2 - t_1$. 
A positive deviation indicates that the upper ball requires more time to cover the same distance, reflecting under-acceleration relative to the lower ball; and a negative deviation implies the opposite, an over-acceleration. The measured deviations for each model are reported in Tab~\ref{tab:double_ball_time_deviation_table}. We find an interesting pattern: the smaller models - Wan 5B and Cosmos 2B show an opposite trend compared to their larger counterparts, with the higher ball accelerating faster than the slower ball. As shown in Tab.~\ref{tab:double_ball_time_deviation_table}. Our gravity adapters (\cref{sec:adapter}) noticeably improve both Wan-5B and Wan-14B, bringing their behavior closer to the physically correct regime in which gravity acts universally, not separately on each object.

\begin{table}[t]
\centering
\caption{\textbf{Time deviation statistics ($\Delta t$ in frames) for the two-ball falling experiment.}
The lower ball hits the ground at time $t_1$ after traversing a vertical distance $y_1$. We then measure the time $t_2$ for the upper ball to traverse the \emph{same} distance and report $\Delta t = t_2 - t_1$ (ideal constant-$g$ motion gives $\Delta t = 0$). Positive values (e.g., Wan~14B, Veo3, Cosmos~14B) mean the upper ball is \emph{slower}, indicating under-acceleration relative to the lower ball; negative values (Wan~5B, Cosmos~2B) mean it is \emph{faster}, indicating over-acceleration. Interestingly, the \emph{direction} of the violation flips with scale: smaller models tend to have negative mean $\Delta t$, while larger models have positive mean $\Delta t$. Gravity Adapters move the means close to zero (e.g., $-0.95$ for 5B and $0.14$ for 14B), thereby shrinking the ranges, which largely cancels this scale-dependent Galileo violation.}
\label{tab:double_ball_time_deviation_table}
\vspace{-5pt}
\setlength{\tabcolsep}{3pt}
\resizebox{0.98\linewidth}{!}{
\begin{tabular}{lccc}
\toprule
\textbf{Model} & \textbf{Mean ($\Delta t$)} & \textbf{Median ($\Delta t$)} & \textbf{Range ($\Delta t$)} \\
\midrule
Cosmos 2B~\cite{nvidia_cosmos_predict2} & -2.77 & -2.0 & [-35.0, 23.0] \\
Cosmos 14B~\cite{nvidia_cosmos_predict2} & 4.03 & 5.0 & [-36.0, 30.0] \\
Veo3~\cite{Google_Veo3_2025} & 6.71 & 7.00 & [-42.0, 28.0] \\
Wan 5B~\cite{wan2025} & -4.22 & -4.0 &  [-35.0, 30.0] \\
Wan 14B~\cite{wan2025} &  2.22 & 1.0 & [-14.0, 16.0] \\
\midrule
Gravity Adapter 5B~\cite{hu2022lora} & -0.95 & -1.00 & [-8.0, 4.0] \\
Gravity Adapter 14B~\cite{hu2022lora} & \textbf{0.14} & \textbf{0.0} & [-5.0, 3.0] \\
\bottomrule
\end{tabular}}
\vspace{-10pt}
\end{table}

\section{Gravity Adapter: Targeted Specialization}
\label{sec:adapter}

Video generators excel as generalists but struggle as physics engines. We investigate whether these physical deficits are due to a lack of model capacity or simply a lack of alignment. By training a lightweight adapter, we probe the learnability of gravitational physics.

\subsection{Training Protocol}
We train a LoRA~\cite{hu2022lora, diffsynth_studio} on Wan~5B to correct its systematic under-acceleration. The training set consists of 100 single-ball drop sequences rendered in diverse HDRI environments distinct from our test benchmark. We train for 5,000 iterations with rank $r = 32$, learning rate $10^{-4}$. Total training time is approximately 6 hours on two A100 GPUs.

\subsection{Results on Benchmark Tasks}

\paragraph{Single-Ball Drops.}
Figure~\ref{fig:og_single_ball_plot} shows that the adapter successfully shifts the mean effective gravity significantly closer to Earth's standard ($1.81 \rightarrow 6.43 m/s^2$). While the distribution remains broad (indicating that stochastic generation artifacts persist), the shift demonstrates that the model can be realigned to physical laws with negligible data (100 examples).
\vspace{-10pt}
\paragraph{Two-Ball Drops (Zero-Shot).}
We also find that the adapter shows zero-shot transfer to two-ball relative timing tasks without ever seeing two-object scenarios during training (Figure~\ref{fig:double_ball_plot}). Timing ratios move substantially closer to the theoretical $t_1^2/t_2^2 = h_1/h_2$ prediction, indicating the adapter learned general gravitational dynamics rather than memorizing single-ball trajectories.

\vspace{-10pt}
\paragraph{Real-World Transfer: PISA Benchmark.}
To test whether the adapter overfits to synthetic training data, we evaluate zero-shot on the PISA benchmark~\cite{li2025pisa}---361 real-world videos of objects falling onto cluttered surfaces. This is challenging: our adapter was trained only on synthetic scenes with sports balls and clear ground planes. Table~\ref{tab:PISA} shows that the adapter  over the baseline across all metrics.

\begin{table}[t]
\centering
\caption{\textbf{Zero-shot generalization to real-world PISA benchmark.} The Gravity Adapter (5B), trained only on 100 synthetic sequences, improves trajectory accuracy on 361 real-world videos~\cite{li2025pisa}. The marginal improvement when full finetuning~(FT) and reward optimization~(ORO)\cite{li2025pisa} does not justify its enormous training cost.}
\label{tab:PISA}
\vspace{-5pt}
\setlength{\tabcolsep}{3pt}
\resizebox{\linewidth}{!}{
\begin{tabular}{l|ccc|c}
\toprule
\textbf{Model} & \textbf{L2} $\downarrow$ & \textbf{Chamfer} $\downarrow$ & \textbf{IoU} $\uparrow$ & \textbf{Params} \\
\midrule

Wan 5B (baseline)~\cite{wan2025} & 0.148 & 0.413 & 0.073 & ---\\
Wan 5B + Gravity Adapter~\cite{hu2022lora} & 0.127 & 0.346 & 0.086 & \textbf{80M} \\
Wan 5B + FT~\cite{li2025pisa} + ORO~\cite{li2025pisa} & \textbf{0.123} & \textbf{0.332} & \textbf{0.089} & 5B \\
\bottomrule
\end{tabular}}
\vspace{-10pt}
\end{table}

\subsection{Comparison with Alternative Approaches}
Table~\ref{tab:additional_guidance} validates our adaptor against three alternatives.

\vspace{-10pt}
\paragraph{LoRA Rank Ablation}
We increase the LoRA rank from 32 to 64 to examine whether additional capacity improves $g_{eff}$. Despite doubling the parameter count (80M $\rightarrow$ 160M), we observe no meaningful improvement, indicating that effective gravity cannot be enhanced through scaling alone. Lower-rank variants (8 and 16) further degrade performance.

\vspace{-10pt}
\paragraph{Explicit Guidance Models}
We test whether providing more information improves physical accuracy. The First-Last Frame model~\cite{wan2025} receives initial and final frames to reduce depth ambiguity, but performs poorly---likely constrained to 5-second generations that default to slow motion. We also evaluate trajectory matching~\cite{wang2025ati} using ground-truth 2D centroid trajectories from drop height to impact, after which we let the model freely generate. This model similarly underperforms ($g_{\mathrm{eff}}=0.38\,\mathrm{m/s^2}$, range $0.04{-}3.22$), suggesting difficulty handling rapid motion. Our adapter exceeds both methods without any explicit guidance.

\begin{figure}[t]
    \centering
    \includegraphics[width=0.9\linewidth]{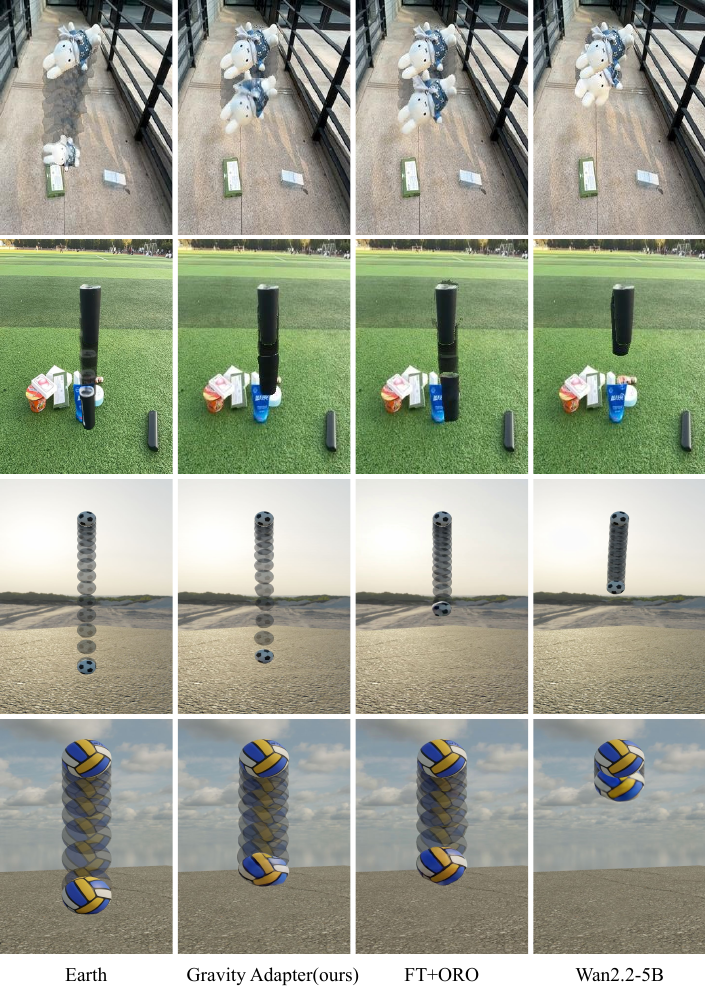}
    \\\vspace{-10pt}
    \caption{\textbf{Comparison with Full FT + ORO.} The Gravity Adapter (5B) shows impressive performance on both the real world dataset~\cite{li2025pisa} and our benchmark with comparable and sometimes better performance at reduced computational cost.
    }
    \label{fig:PISA}
    \vspace{-10pt}
\end{figure}

\vspace{-10pt}
\paragraph{Full Model Fine-Tuning}
Replicating PISA's protocol~\cite{li2025pisa}, we fine-tune all Wan~5B parameters on our 100-example set using their two-stage approach: 5,000 supervised iterations followed by 1,000 reward-model iterations~\cite{ravi2024sam2}. Due to computational constraints, we reduce resolution to 480p and 10 denoising steps. Our LoRA matches or exceeds full fine-tuning performance (Fig.~\ref{fig:PISA}) while training only 1\% of parameters and using an order of magnitude less compute, suggesting physics correction may be amenable to low-rank adaptation.

\begin{table}[t]
\centering
\caption{The Gravity Adapter (5B) exceeds full fine-tuning performance on our benchmark with less training computation and outperforms methods with additional guidance.}
\label{tab:additional_guidance}
\vspace{-5pt}
\setlength{\tabcolsep}{3pt}
\resizebox{0.98\linewidth}{!}{
\begin{tabular}{lccc}
\toprule
\textbf{Method} & \textbf{Mean ($\mathrm{m/s^2}$)} & \textbf{Median ($\mathrm{m/s^2}$)} & \textbf{Range ($\mathrm{m/s^2}$)} \\
\midrule
Wan~5B (baseline)~\cite{wan2025} & 1.81 & 1.24 &  [0.26, 8.26] \\
Baseline + FT~\cite{li2025pisa} + ORO~\cite{li2025pisa}  & 4.07 & 3.63& [0.89, 10.72] \\
Gravity Adapter (ours, rank 8)~\cite{hu2022lora} & 3.07 & 2.83& [1.24, 16.64] \\
Gravity Adapter (ours, rank 16)~\cite{hu2022lora} & 5.64 & 5.72 & [1.51, 12.05] \\
Gravity Adapter (ours, rank 32)~\cite{hu2022lora} & \textbf{6.43} & \textbf{6.38} & [1.24, 16.64] \\
Gravity Adapter (ours, rank 64)~\cite{hu2022lora} & 6.11 & 5.72 & [1.91, 14.17] \\
\midrule
FLF (first + last frame)~\cite{wan2025} & 0.58 & 0.22 & [0.03, 23.70] \\
Trajectory guided~\cite{wang2025ati} & 0.38& 0.16 & [0.04, 3.22] \\
\bottomrule
\end{tabular}}
\vspace{-10pt}
\end{table}

\subsection{Generalization to Canonical Scenarios}

\begin{figure}[t]
    \centering
    \includegraphics[width=0.9\linewidth]{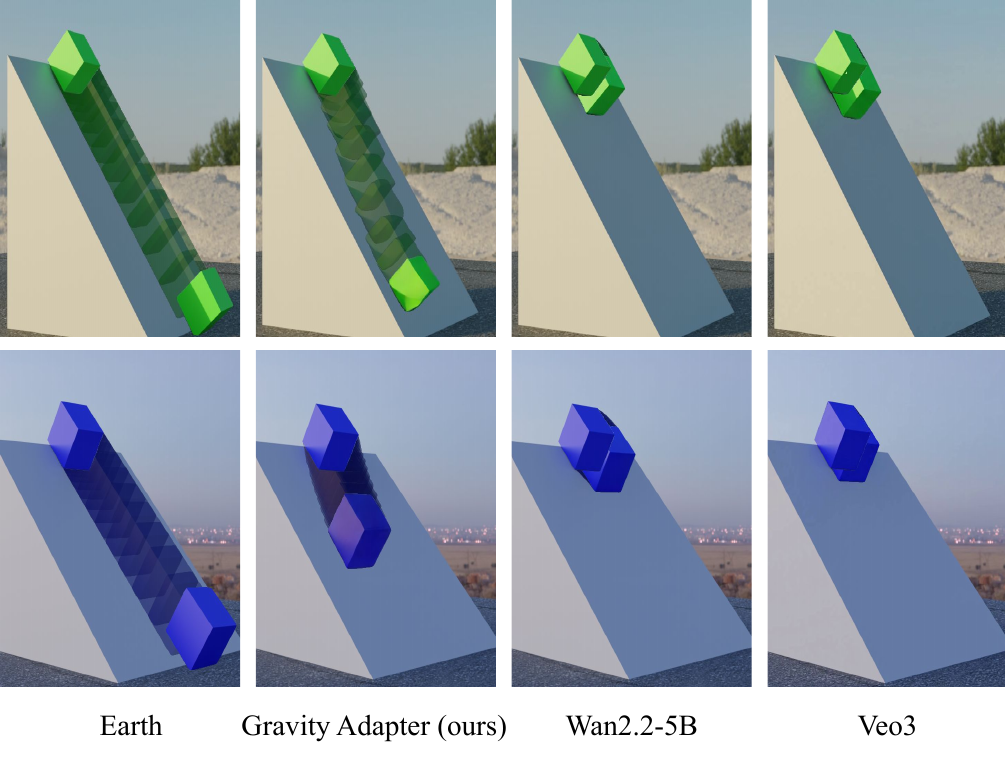}
    \\\vspace{-10pt}
    \caption{\textbf{Zero-shot generalization to inclined planes.} The Gravity Adapter (5B) improves motion realism for cubes sliding down inclines at various angles, despite never seeing this scenario during training. This demonstrates learning of general gravitational principles rather than ball-specific heuristics.}
    \vspace{-10pt}
    \label{fig:incline}
\end{figure}

To test whether the adapter learned general gravitational principles or merely ball-dropping heuristics, we evaluate on canonical physics regimes never seen during training:

\vspace{-10pt}
\paragraph{Inclined Planes.}
We generate 12 sequences of smooth cubes sliding down frictionless inclines at angles from $30^\circ$ to $75^\circ$. Under correct physics, acceleration should scale as $g\sin(\theta)$. Figure~\ref{fig:incline} shows the Gravity Adapter substantially improves motion realism compared to baseline Wan~5B, with objects accelerating appropriately for their incline angle. This demonstrates that the adapter internalized gravitational principles applicable beyond vertical free fall.

\section{Discussion}
\label{sec:conclusion}

We evaluate video generators on gravitational physics, revealing systematic failures in their native time gauge. Single-ball experiments show all models dramatically under-accelerate ($g_{\mathrm{eff}}\!\sim\!1\text{--}2\,\mathrm{m/s^2}$ vs.\ Earth's $9.81\,\mathrm{m/s^2}$), so balls fall much more slowly than real-world gravity would predict, exhibiting an effective ``sub-Earth'' gravity. A simple global temporal rescaling can move the \emph{mean} $g_{\mathrm{eff}}$ closer to $9.81\,\mathrm{m/s^2}$, but the resulting distributions remain broad with large variance and heavy tails. Our unit-free two-ball protocol eliminates scale ambiguity by testing timing ratios ($t_1^2/t_2^2 = h_1/h_2$) and time deviations $\Delta t$ for traversing the same distance. Most models catastrophically fail Galileo's principle, with different objects effectively experiencing different gravity within the same scene; these relative-timing violations are invariant to any global time rescaling. Surprisingly, larger models perform worse on these metrics, contradicting naive scaling-law expectations. A lightweight LoRA adapter trained on 100 synthetic examples substantially corrects these failures (e.g., $1.81 \rightarrow 6.43\,\mathrm{m/s^2}$) and moves $\Delta t$ close to zero with tighter ranges, and it generalizes zero-shot to real-world videos, two-ball drops, and inclined planes.

\vspace{-10pt}
\paragraph{Why Focus on Gravity Alone?}
We deliberately constrain our investigation to gravity for three reasons. First, gravity is the most fundamental physical law. If models fail to account for constant acceleration, they cannot handle projectile motion, pendulums, or complex dynamics. Second, gravity provides unambiguous ground truth: the relationship $t \propto \sqrt{h}$ is mathematically exact, unlike friction coefficients or collision elasticity, which depend on materials. Third, our unit-free protocol exploits gravity's universality -- the ratio $t_1^2/t_2^2 = h_1/h_2$ cancels scale, focal length, frame rate, and the absolute value of $g$. This approach establishes a methodological template for evaluating other physical laws.

\vspace{-10pt}
\paragraph{Appearance Models vs World Models.}
Video generators produce visually plausible motion without obeying simple physical laws in our benchmark. They learn correlational patterns, not causal mechanics. A world model must predict both appearance and causality; current generators only reliably achieve the former. In their default parameterization, these systems act as if objects fall at $\sim\!20\%$ of Earth's gravity and as if different objects can experience different accelerations within the same scene. Systems with such behavior cannot reliably simulate environments for robotics, autonomous vehicles, or scientific visualization. Our adapter's preliminary success suggests video models may encode latent physical structure that could benefit from targeted activation rather than learning from scratch or scaling up.

\vspace{-10pt}
\paragraph{Why Under-Acceleration?}
We hypothesize four contributing factors. {(1) Training data:} Video datasets contain slow-motion and time-lapse footage, biasing models toward non-physical speeds. {(2) Temporal architecture:} Frame downsampling and discrete tokenization smooth motion profiles, eliminating fine-grained acceleration. {(3) Reward signal:} Perceptual losses optimize visual quality without penalizing incorrect dynamics -- slow falling ``looks plausible'' despite wrong physics, and alignment procedures (e.g., RLHF) likely prioritize aesthetics over dynamics. {(4) Compositional reasoning:} Models learn state transitions (``ball at height'' $\rightarrow$ ``ball descending'') rather than continuous dynamics $h(t) = h_0 - \tfrac{1}{2}gt^2$. The adapter's 100-example effectiveness suggests the primary issue is incorrect weighting of existing knowledge rather than a lack of representational capacity.

\vspace{-10pt}
\paragraph{The Scaling Paradox.}
Larger models show worse gravity behavior, contradicting standard expectations that more capacity and data should monotonically improve physical fidelity. Both single-ball effective gravity and two-ball $\Delta t$ reveal that scaling changes the \emph{bias} rather than eliminating the violation: smaller models (e.g., Cosmos~2B, Wan~5B) tend to over-accelerate the higher ball (negative mean $\Delta t$), while larger models (Cosmos~14B, Wan~14B) tend to under-accelerate it (positive mean $\Delta t$). Scale appears to improve visual abstraction, not physical simulation. Alignment procedures (RLHF) likely optimize aesthetic preferences rather than physical accuracy. %

\vspace{-10pt}
\paragraph{Limitations and Future Work.}
Our adapter achieves only 65\% of terrestrial gravity on average with high variance ($1.24\text{--}16.64\,\mathrm{m/s^2}$). We evaluate only vertical free-fall and simple inclined motion, not projectile trajectories, rotation, collisions, or friction. Future directions include: (1) {Scaling training data:} Test whether $1{,}000+$ examples achieve full accuracy and reduced variance. (2) {Physics-informed losses:} Incorporate $d = \tfrac{1}{2}gt^2$ and relative-timing penalties directly into training objectives. (3) {Multi-law adapters:} Train single adapters for multiple physical phenomena. (4) {Hybrid architectures:} Integrate explicit physics simulation modules. (5) {Broader evaluation:} Apply unit-free protocols to momentum, energy conservation, and fluid dynamics.

{
    \small
    \bibliographystyle{ieeenat_fullname}
    \bibliography{main}
}

\clearpage
\maketitlesupplementary

\begin{figure*}[t]
\centering

\begin{subfigure}{0.23\textwidth}
    \includegraphics[width=\linewidth]{figures/og_single_ball_plot.png}
    \caption{Original}
    \label{fig:supple_og_single_ball_plot}
\end{subfigure}
\hfill
\begin{subfigure}{0.23\textwidth}
    \includegraphics[width=\linewidth]{figures/time_scaled_single_ball_plot.png}
    \caption{Mean Scaling}
    \label{fig:supple_time_scaled_single_ball_plot}
\end{subfigure}
\hfill
\begin{subfigure}{0.23\textwidth}
    \includegraphics[width=\linewidth]{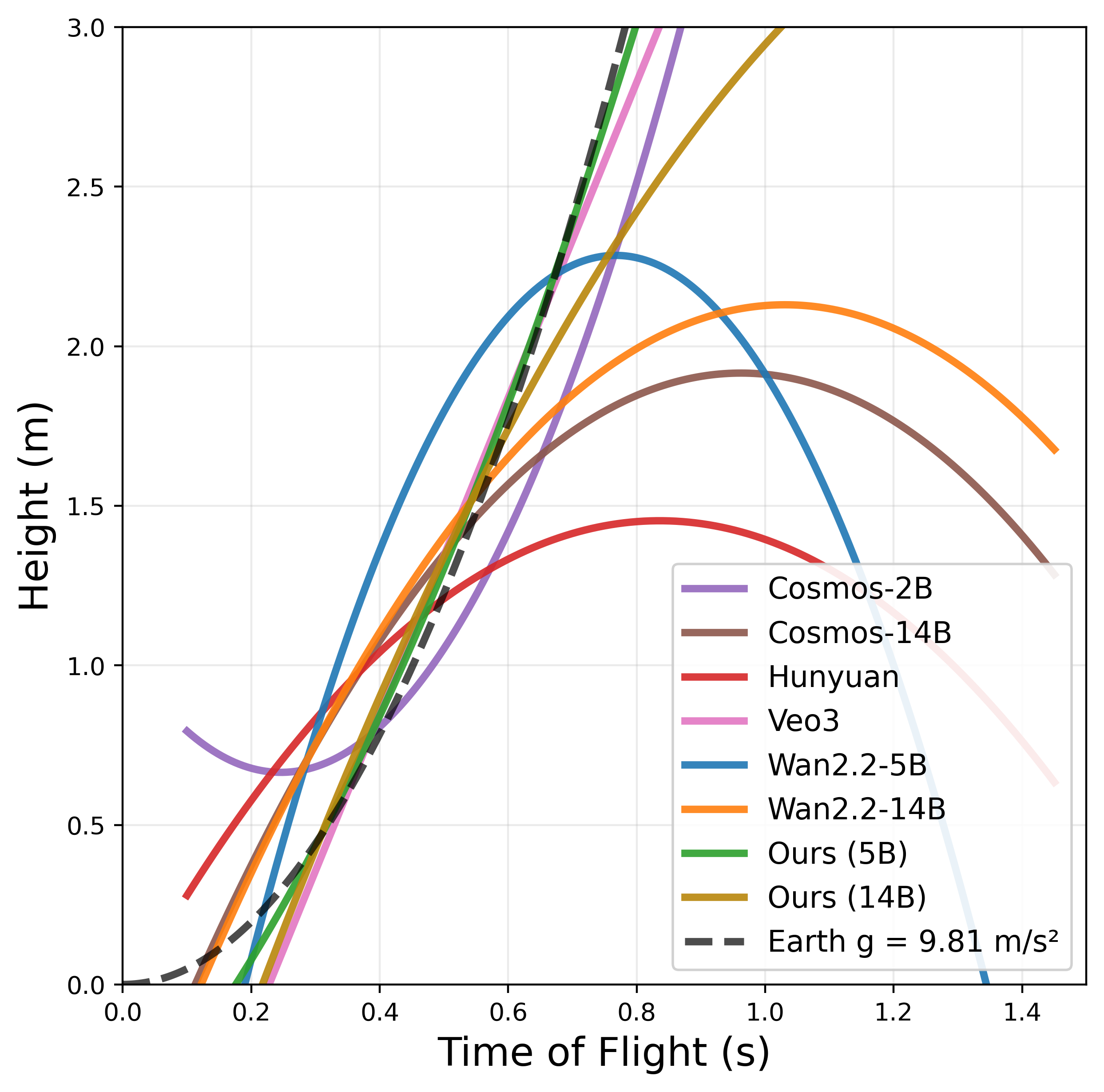}
    \caption{Per-Sample Scaling}
    \label{fig:supple_per_sample_time_scaled_single_ball_plot}
\end{subfigure}
\hfill
\begin{subfigure}{0.23\textwidth}
    \includegraphics[width=\linewidth]{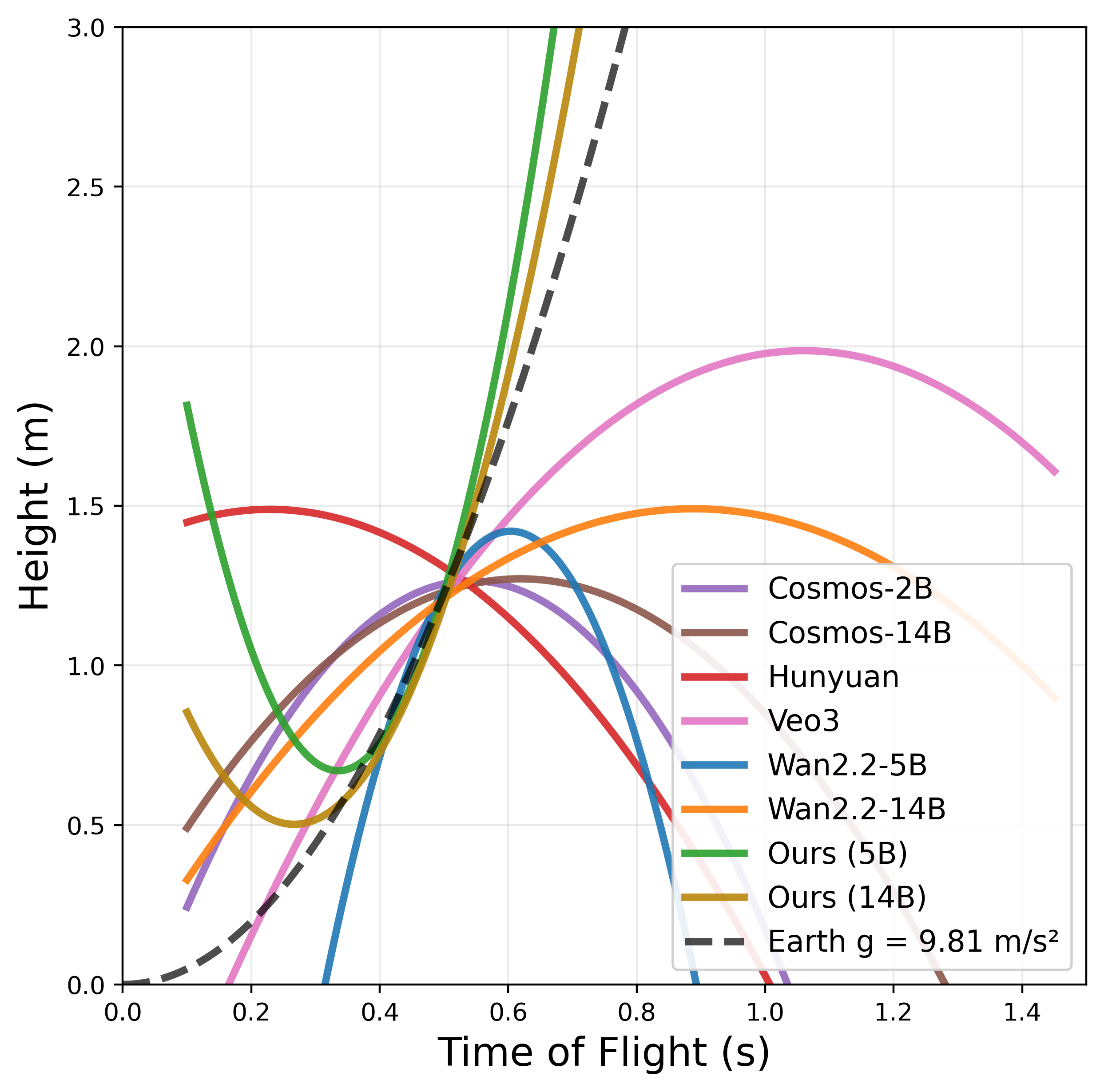}
    \caption{Height-Adjusted Time Scaling}
    \label{fig:supple_time_scaled_height_adj_single_ball_plot}
\end{subfigure}

\caption{
\textbf{Effect of time-scaling on $h$–$t$ relationships.}
\textbf{(a)} We plot $h$ versus $t$ for all models. We repeat each test example with 4 seeds and fit polynomials through the means. The gray dashed line indicates terrestrial motion. All models systematically under-accelerate, and none obey the square root scaling law of time with height. The Gravity Adapters (green, gold) substantially improves Wan 5B and Wan 14B towards correct gravity.
\textbf{(b) Mean time scaling.} We compute a Mean time scalar using a subset of random 30 samples from our dataset, which scales the effective time of the 30 samples to better match the ground truth time. The Mean time scalar, when applied to the second subset of 45 samples, brings the mean effective gravity close to 9.81, m/s2 for many models.
\textbf{(c) Per-sample scaling.} Instead of a mean time scalar, we experiment with a per sample time scalar, computed using half of the seeds. The scalar is then applied to the other half of the seeds.
\textbf{(d) Height-adjusted Mean Time scaling.} Accounting for deviation from the straight line vertical trajectory does not significantly improve the effective gravity.
}
\label{fig:supple_single_ball_plot_grid}
\end{figure*}

\begin{figure*}[t]
    \centering
    \begin{subfigure}[b]{0.48\textwidth}
        \centering
        \includegraphics[width=\textwidth]{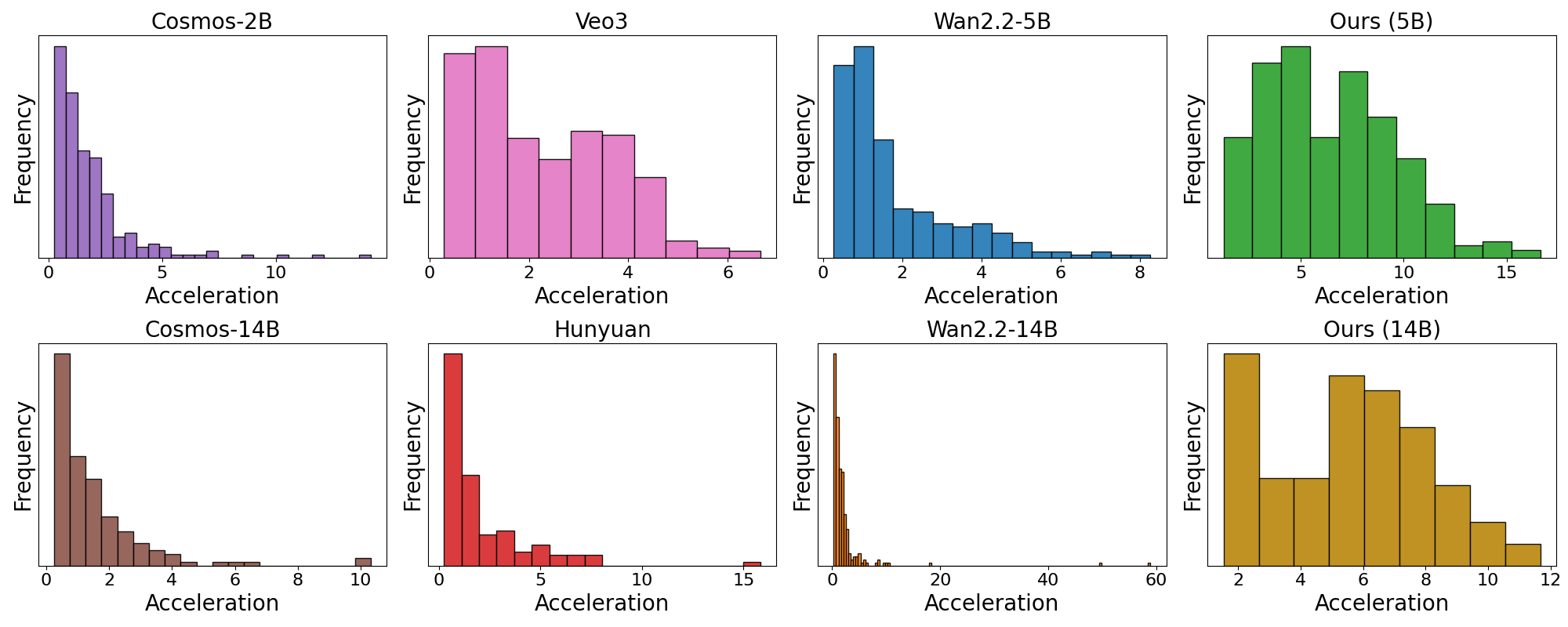}
        \caption{Original}
        \label{fig:og_single_ball_histogram}
    \end{subfigure}
    \hfill
    \begin{subfigure}[b]{0.48\textwidth}
        \centering
        \includegraphics[width=\textwidth]{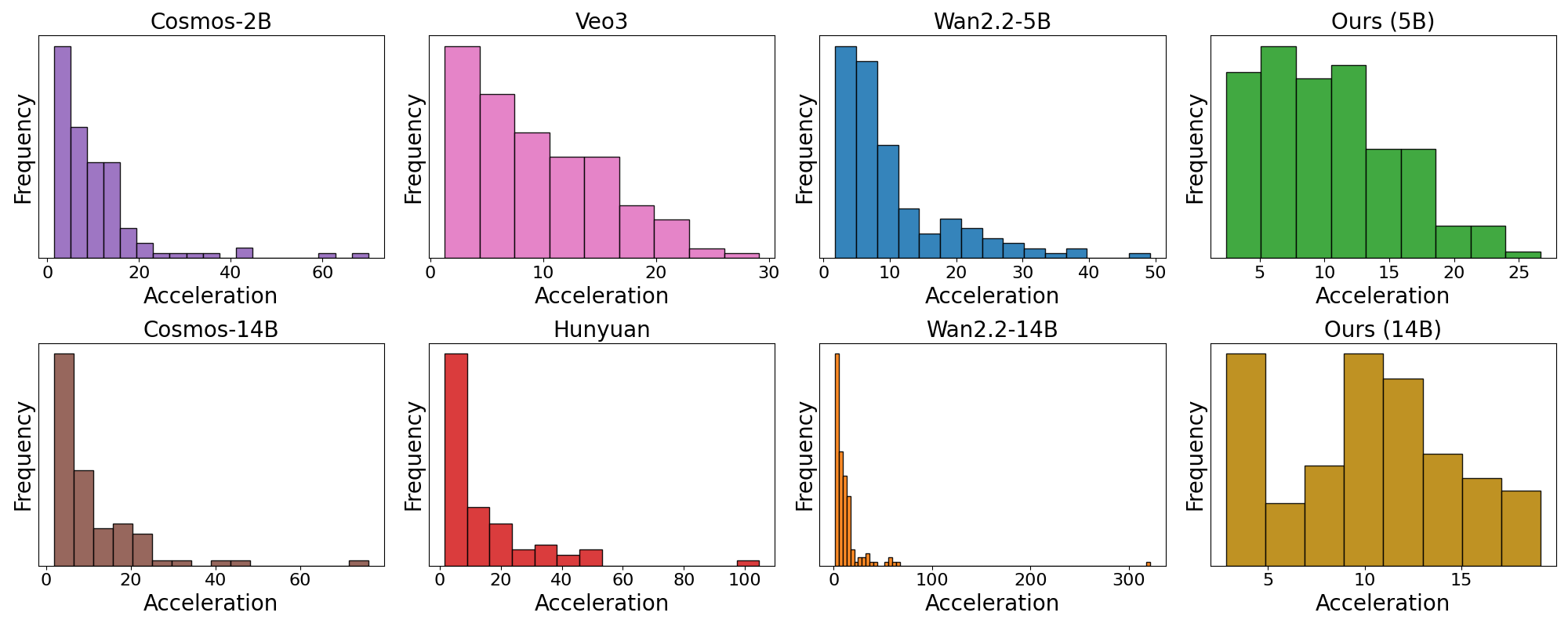}
        \caption{Mean Time Scaling}
        \label{fig:time_scaled_single_ball_histogram}
    \end{subfigure}

    \vspace{8pt}

    \begin{subfigure}[b]{0.48\textwidth}
        \centering
        \includegraphics[width=\textwidth]{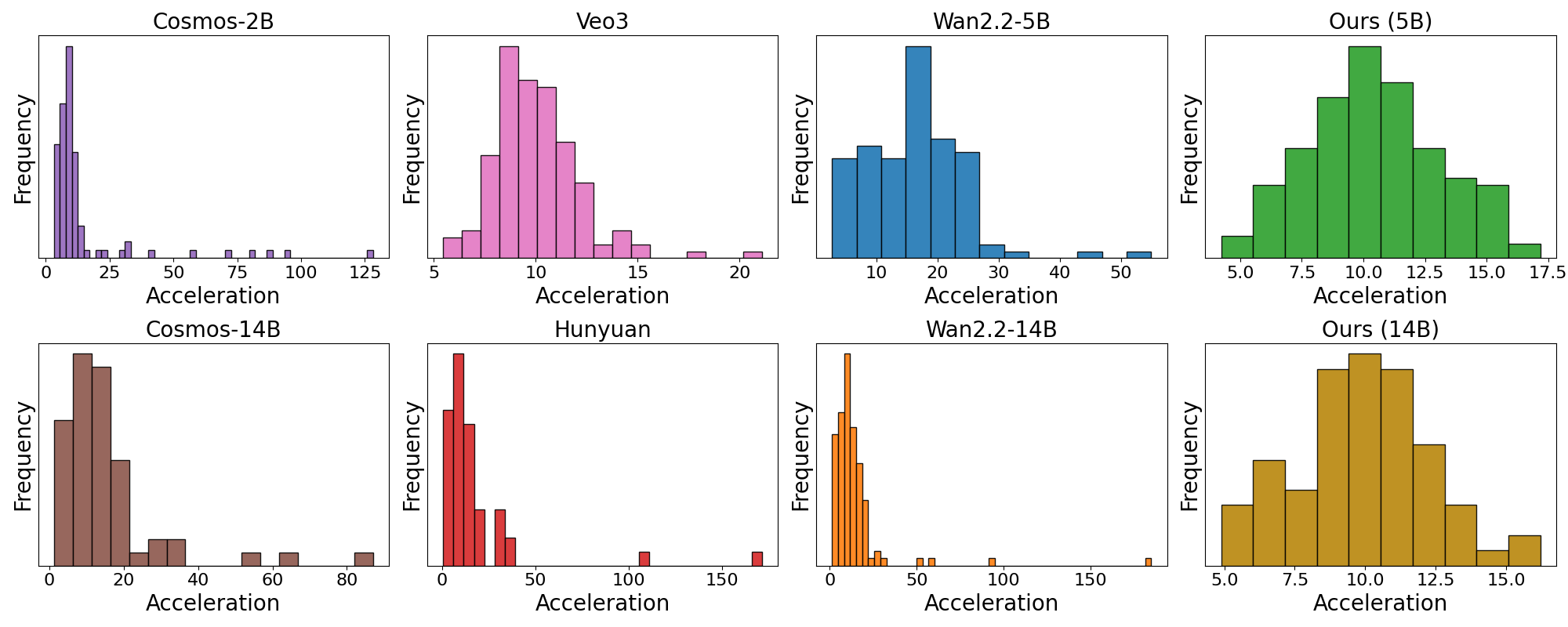}
        \caption{Per sample Time Scaling}
        \label{fig:per_sample_time_scaled_single_ball_histogram}
    \end{subfigure}
    \hfill
    \begin{subfigure}[b]{0.48\textwidth}
        \centering
        \includegraphics[width=\textwidth]{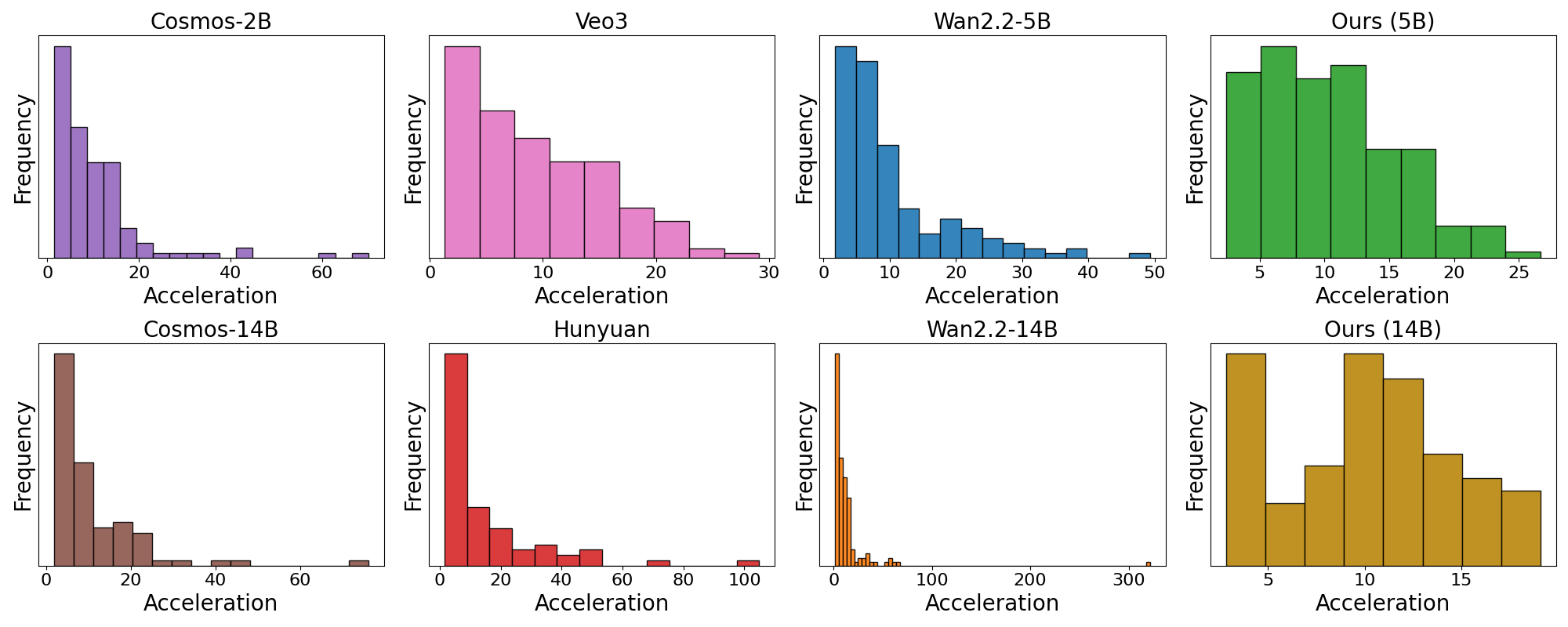}
        \caption{Mean Time Scaling with Height adjustments}
        \label{fig:time_scaled_height_adj_single_ball_histogram}
    \end{subfigure}

    \caption{\textbf{Effect of Time scaling on distribution of gravities}} %
    \label{fig:accel_histograms}
\end{figure*}

\section{Additional Time Scaling Heuristics}
\textbf{Per Sample Time Scaling}
We also evaluate per-sample time scaling to account for seed-to-seed variance. For each sample, we use seeds 999 and 777 to compute an individual scaling factor, then apply it to seeds 42 and 123. Per-sample scaling yields higher $g_{scaled}$ values (Tab.~\ref{tab:supple_per_sample_time_scaled_single_ball_table}) but the variance remains substantial, possibly due to high variance between seeds. 

\textbf{Adjusting height for angled trajectories.}
Many generated videos show non-vertical trajectories, which could result from the model interpreting the ground plane or camera as tilted. To provide the benefit of the doubt again, we adjust the effective drop height based on the observed angle of deviation, computing $h_{adj}$ and the corresponding scaled gravity $g_{adj+scaled}$ using globally scaled times. Tab.~\ref{tab:supple_time_scaled_height_adj_single_ball_table} shows that this height adjustment provides no meaningful improvement. Even after applying perspective correction, the models still show substantial under-acceleration. The gravity distributions (Fig.~\ref{fig:time_scaled_single_ball_histogram}--\ref{fig:time_scaled_height_adj_single_ball_histogram}) reveal that most samples remain below $9.81,\mathrm{m/s^2}$, indicating that time scaling alone does not resolve the core physical error. Fig.~\ref{fig:supple_time_scaled_single_ball_plot}--\ref{fig:supple_time_scaled_height_adj_single_ball_plot} shows the change in the $h$ vs $t$ plots when different time scaling techniques are applied. 

\section{Prompt Structure}
\label{sec:prompt_structure}

We use a consistent prompt structure across all experiments to isolate physics understanding from prompt sensitivity.
\vspace{-1pt}
\paragraph{Single Ball Drops.} A video showing a ball being dropped from a height onto the ground. The camera is static and positioned to clearly capture the vertical motion of the ball. The ball falls naturally under gravity, accelerating freely with no air resistance and hits the ground. 
\vspace{-1pt}
\paragraph{Two Ball Drops.}
A video showing two identical balls being dropped from two different heights onto the ground. The camera is static and positioned to clearly capture the vertical motion of both balls. Both balls fall naturally under gravity, accelerating freely with no air resistance, and hit the ground. 
\vspace{-1pt}
\paragraph{Expanded Prompt.} A video showing a ball of diameter $diameter$ meters being dropped from a height of $initial$ $height$ meters onto the ground. The camera is static and positioned at a height of $camera$ $height$ meters, at a distance of $camera$ $depth$ meters to clearly capture the vertical motion of the ball. The ball falls naturally under gravity at $9.8$ $meters$ $per$ $second$ $square$, accelerating freely with no air resistance, and hits the ground. 
\vspace{-1pt}
\paragraph{Inclined Plane.} A video showing a smooth square block sliding down a frictionless inclined plane under gravity. The inclined plane is fixed and set at an angle, with no friction between the block and the surface. The block starts from rest near the top and accelerates uniformly as it slides downward. The camera is static and positioned at the side to clearly capture the motion along the incline.

{
\setlength{\tabcolsep}{2pt}
\begin{table*}[t!]
\caption{\textbf{Effect of time-scaling on $g_{\mathrm{eff}}$ estimation across models.}
\textbf{(a) Original.} We report effective gravity values computed as 
$g_{\mathrm{eff}} = 2h/t^2$ (m/s$^2$). The ground truth is $9.81$ m/s$^2$. 
All models under-accelerate, and Gravity Adapters consistently reduce this deficit. 
Reported mean values are averaged over four random seeds and all test examples. Median and Range values are across all seeds and test samples.
\textbf{(b) Mean time scaling.} A global scalar(MTS) is estimated from a 
30-sample subset and applied to a disjoint 45-sample split. This shifts several models 
closer to $9.81$ m/s$^2$, but variance remains high. 
\textbf{(c) Per-sample time scaling.} Instead of a global correction, we compute 
a per-sample scalar using half the seed runs and apply it to the others. 
\textbf{(d) Height-adjusted scaling.} Accounting for deviations from ideal 
vertical motion does not meaningfully improve $g_{\mathrm{eff}}$ estimation, suggesting 
that model errors stem from deeper physics issues rather than trajectory shape.}
\centering
\begin{subtable}{0.95\columnwidth}
\centering
\caption{Original}
\label{tab:supple_og_single_ball_table}
\tiny
\begin{tabularx}{0.95\linewidth}{lcccc}
\toprule
\textbf{Model} & \textbf{Mean ($\mathrm{m/s^2}$)} & \textbf{Median ($\mathrm{m/s^2}$)}  & \textbf{Range ($\mathrm{m/s^2}$)} &{\textbf{Q1-Q3} ($\mathrm{m/s^2}$)} \\
\midrule
 Cosmos~2B~\cite{nvidia_cosmos_predict2}&  \textbf{1.85} & 1.30 &  [0.23, 14.18] &[0.68, 2.24] \\
Cosmos~14B~\cite{nvidia_cosmos_predict2}&  1.51&  1.01&  [0.24,10.31] &[0.55, 1.92]  \\
 Hunyuan~\cite{kong2024hunyuanvideo}&  1.97&  1.15&   [0.23,15.84] &[0.48, 2.72]\\
 Veo3~\cite{Google_Veo3_2025}&  2.27 &  2.08& [0.28, 6.66] &[1.00, 3.38]\\
 Wan~5B~\cite{wan2025}&  1.81&  \textbf{1.24} &   [0.26,8.26] &[0.77, 2.41]\\
 Wan~14B~\cite{wan2025}&  2.18& 1.19 & [0.27, 59.98] & [0.63, 2.04]    \\
 \midrule
Gravity Adapter~5B~\cite{hu2022lora}&  \textbf{6.43} & 6.38 & [1.24, 16.64] &[3.95, 8.65]\\
Gravity Adapter~14B~\cite{hu2022lora} &  \textbf{5.51} &  5.63&  [1.52, 11.67] &[3.02, 7.45] \\
\bottomrule
\end{tabularx}
\end{subtable}
\hfill
\begin{subtable}{0.95\columnwidth}
\centering
\caption{Mean-Time Scaled}
\label{tab:supple_time_scaled_single_ball_table}
\tiny
\begin{tabularx}{0.98\linewidth}{lccccc}
\toprule
\textbf{Model} & \textbf{MTS} & \textbf{Mean}($\mathrm{m/s^2}$) & \textbf{Median} ($\mathrm{m/s^2}$) & \textbf{Range} ($\mathrm{m/s^2}$) &{\textbf{Q1-Q3} ($\mathrm{m/s^2}$)}\\
\midrule
  Cosmos~2B~\cite{nvidia_cosmos_predict2}&  2.43& 10.34 &  7.24 & [1.40, 69.96] &[3.74, 12.75]  \\
 Cosmos~14B~\cite{nvidia_cosmos_predict2}&  2.77 & 10.86 & 7.19 & [1.81, 76.04] &[4.02, 14.20] \\
  Hunyuan~\cite{kong2024hunyuanvideo}&  2.63 & 13.85 & 7.06 & [1.55, 104.64] &[3.23, 18.93] \\
  Veo3~\cite{Google_Veo3_2025}& 2.12 & 9.39 & 8.5 & [1.26, 29.11]  &[4.17, 13.69] \\
  Wan~5B~\cite{wan2025}& 2.43 & 10.24 & 7.22 & [1.76, 49.15] &[4.60, 13.05]\\
  Wan~14B~\cite{wan2025}& 2.56 & 13.78 & 7.78 & [1.75, 321.30]  &[4.11, 14.11]\\
  \midrule
Gravity Adapter~5B~\cite{hu2022lora}& 1.28 & 10.27 & 9.74 & [2.4, 26.67] &[5.98, 13.44]\\
Gravity Adapter~14B~\cite{hu2022lora}& 1.36 & 10.00 & 10.10 & [2.82, 19.08]  &[5.64, 13.01]\\
\bottomrule
\end{tabularx}
\end{subtable}
\vspace{2.5em} 
\begin{subtable}{0.95\columnwidth}
\centering
\caption{Per Sample Time Scaled}
\label{tab:supple_per_sample_time_scaled_single_ball_table}
\tiny
\begin{tabularx}{0.95\linewidth}{lcccc}
\toprule
\textbf{Model} & \textbf{Mean ($\mathrm{m/s^2}$)} & \textbf{Median ($\mathrm{m/s^2}$)}  & \textbf{Range ($\mathrm{m/s^2}$)} &{\textbf{Q1-Q3} ($\mathrm{m/s^2}$)}\\
\midrule
 Cosmos~2B~\cite{nvidia_cosmos_predict2}&  15.08& 8.69 &  [3.23, 128.26] &[6.57, 11.78]  \\
Cosmos~14B~\cite{nvidia_cosmos_predict2}&  15.23&  11.75&  [1.26, 87.06]  &[6.95, 16.80]\\
 Hunyuan~\cite{kong2024hunyuanvideo}&  17.95&  10.79&   [0.64, 171.55]&[6.80, 16.93]\\
 Veo3~\cite{Google_Veo3_2025}&  10.04 &  9.77 &  [5.45, 21.10] &[8.60, 11.10]\\
 Wan~5B~\cite{wan2025}&  16.31&  16.98 &   [2.77, 54.95] &[9.71, 20.04]\\
 Wan~14B~\cite{wan2025}&  14.05 & 10.77 & [1.43, 184.81]    &[7.11, 15.64]\\
 \midrule
Gravity Adapter~5B~\cite{hu2022lora}&  10.46 & 10.24 & [4.23, 17.17]&[8.62, 12.11]\\
Gravity Adapter~14B~\cite{hu2022lora} &  9.88 &  9.81 &  [4.88, 16.21] &[8.61, 11.07]\\
\bottomrule
\end{tabularx}
\end{subtable}
\hfill
\begin{subtable}{0.95\columnwidth}
\centering
\caption{Mean-Time Scaled with Height Adjustments}
\label{tab:supple_time_scaled_height_adj_single_ball_table}
\tiny
\begin{tabularx}{0.98\linewidth}{lccccc}
\toprule
\textbf{Model} & \textbf{MTS} & \textbf{Mean}($\mathrm{m/s^2}$) & \textbf{Median} ($\mathrm{m/s^2}$) & \textbf{Range} ($\mathrm{m/s^2}$) &{\textbf{Q1-Q3} ($\mathrm{m/s^2}$)}\\
\midrule
  Cosmos~2B~\cite{nvidia_cosmos_predict2}&  2.43& 10.35 &  7.24 & [1.40, 70.06] &[3.74, 12.77] \\
 Cosmos~14B~\cite{nvidia_cosmos_predict2}&  2.77 & 10.89 & 7.21 & [1.81, 76.12]&[4.02, 14.21] \\
  Hunyuan~\cite{kong2024hunyuanvideo}&  2.63 & 14.59 & 7.07 & [1.55, 104.82] &[3.23, 19.71]\\
  Veo3~\cite{Google_Veo3_2025}& 2.12 & 9.39 & 8.51 & [1.3, 29.12] &[4.17, 13.69]\\
  Wan~5B~\cite{wan2025}& 2.43 & 10.25 & 7.22 & [1.76, 49.29] &[4.60, 13.05]\\
  Wan~14B~\cite{wan2025}& 2.56 & 13.86 & 7.78 & [1.76, 321.56]  &[4.11, 14.11]\\
  \midrule
Gravity Adapter~5B~\cite{hu2022lora}& 1.28 & 10.27 & 9.74 & [2.4, 26.67] &[5.98, 13.44]\\
Gravity Adapter~14B~\cite{hu2022lora}& 1.36 & 10.00 & 10.13 & [2.82, 19.08]  &[5.64, 13.01]\\
\bottomrule
\end{tabularx}
\end{subtable}
\label{fig:supple_single_ball_table_grid}
\end{table*}
}

\begin{figure*}[t]
  \centering
  \includegraphics[width=0.7\linewidth,keepaspectratio]{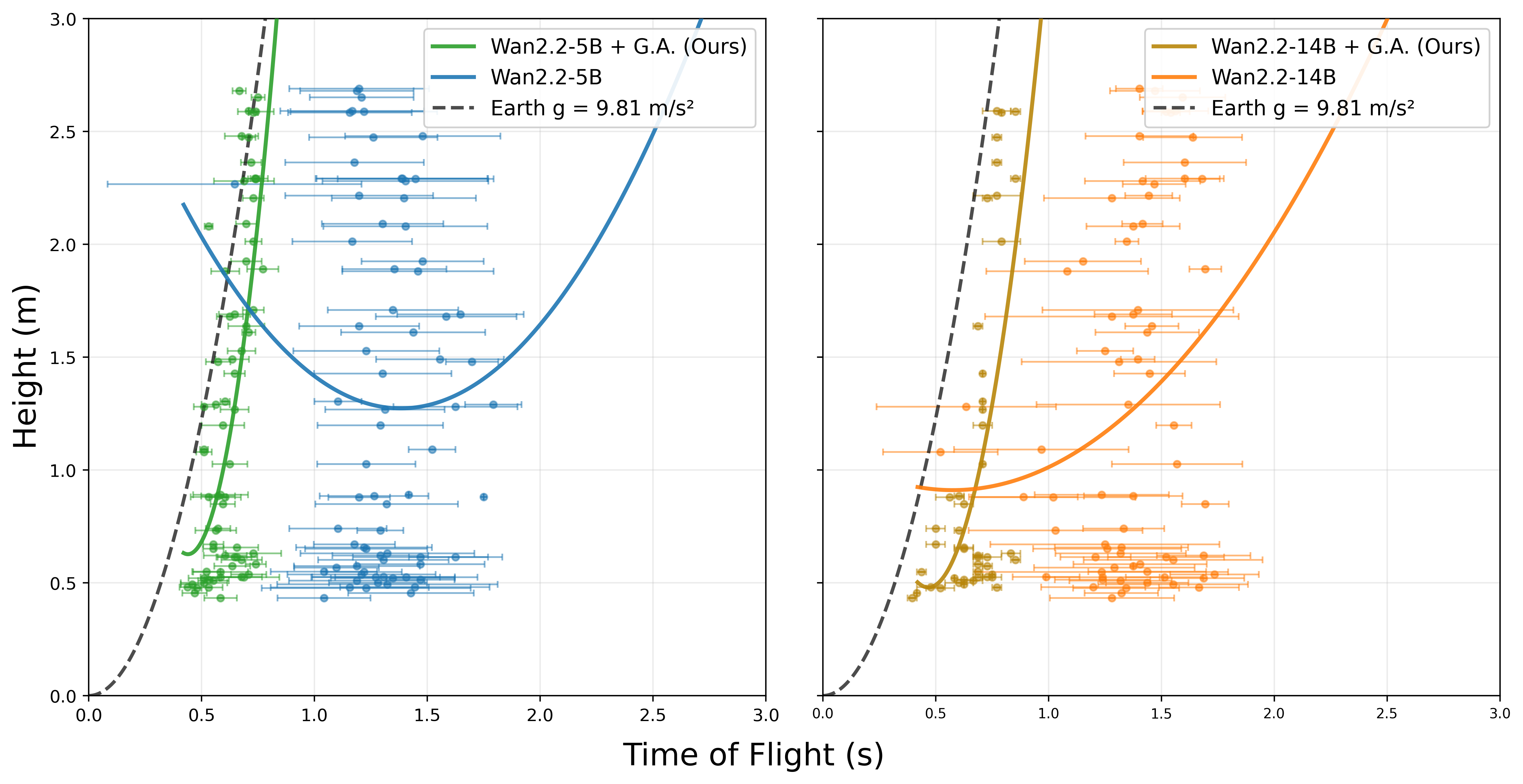}
  \vspace{-10pt}
  \caption{\textbf{Single ball falling results for gravity adapters.} We plot $h$ versus $t$ values for Wan~5B and Wan~14B with and without gravity adapters. The black dashed line represents terrestrial motion. We scatter plot the mean of each sample across 4 seeds and show error bars. The gravity adapters improve the base model to correct gravity while minimizing variance across seeds.
  }
  \label{fig:lora_adapters_single_ball}
  \vspace{-10pt}
\end{figure*}

\begin{figure*}[t]
  \centering
  \includegraphics[width=0.7\textwidth,keepaspectratio]{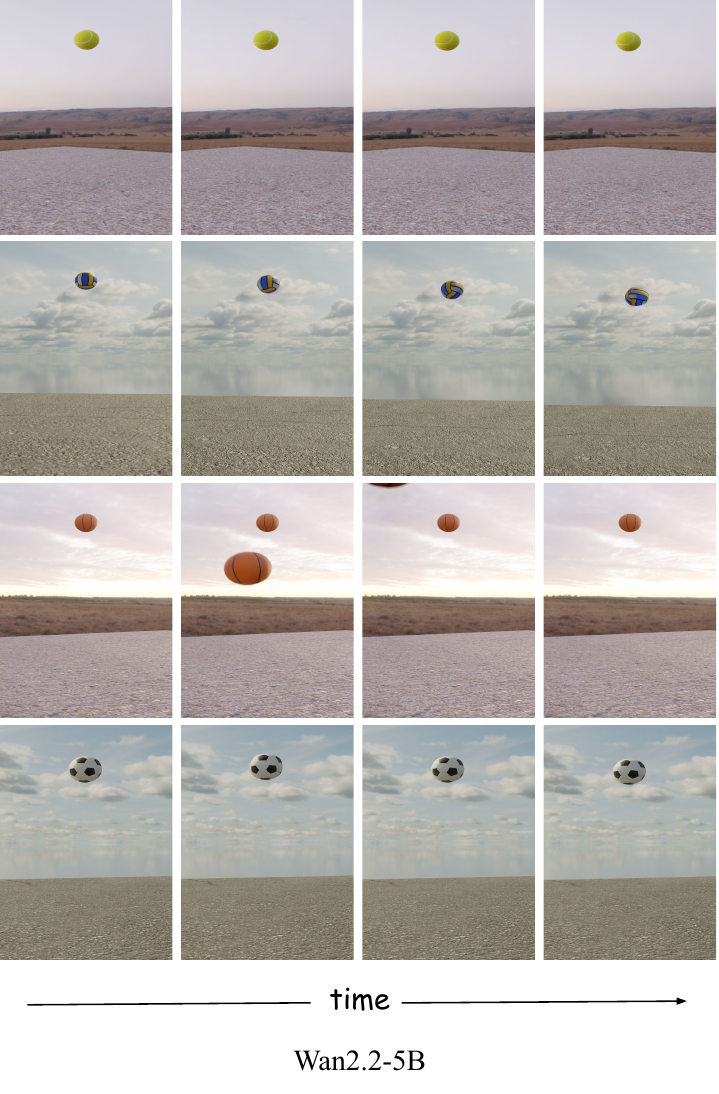}
  \vspace{-10pt}
  \caption{\textbf{Ablation on number of frames generated.} We observe that reducing the generation window to 1 second leads to poor performance. We show representative examples where either the ball remains suspended rather than falling, or the model generates an unintended extra object. Due to this, we set the generation time to 2 seconds.
  }
  \label{fig:supple_1sec}
  \vspace{-10pt}
\end{figure*}

\begin{figure*}[t]
  \centering
  \includegraphics[width=0.7\textwidth,keepaspectratio]{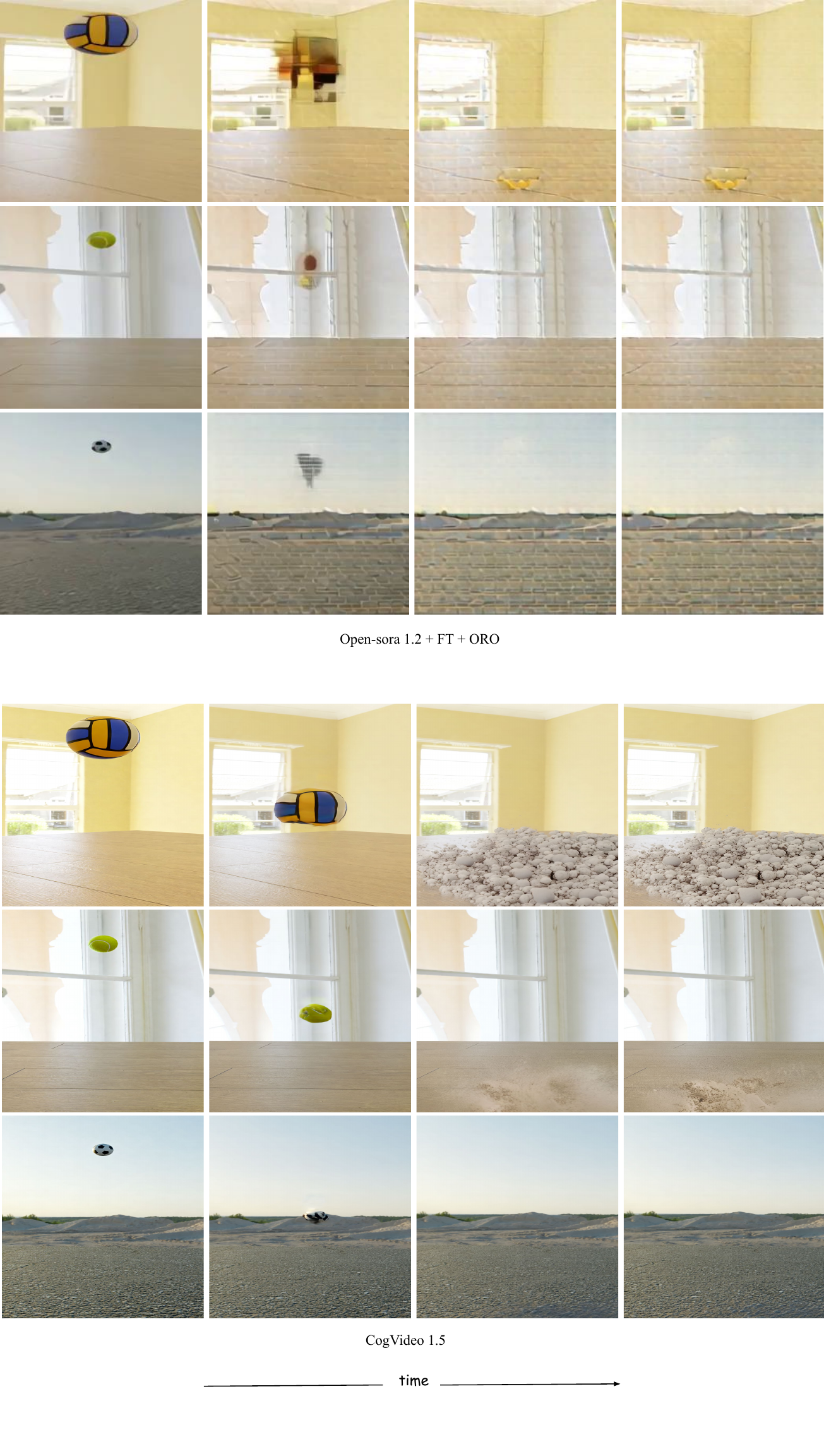}
  \vspace{-10pt}
  \caption{\textbf{Additional models tested.} We also generate samples using OpenSora finetuned in the style of \cite{li2025pisa} and CogVideoX-1.5 \cite{yang2024cogvideox}. However, the resulting videos exhibit severe artifacts and hallucinations, making them unsuitable for meaningful evaluation..
  }
  \label{fig:supple_other_models}
  \vspace{-10pt}
\end{figure*}

\begin{figure*}[t]
  \centering
  \includegraphics[width=\textwidth,keepaspectratio]{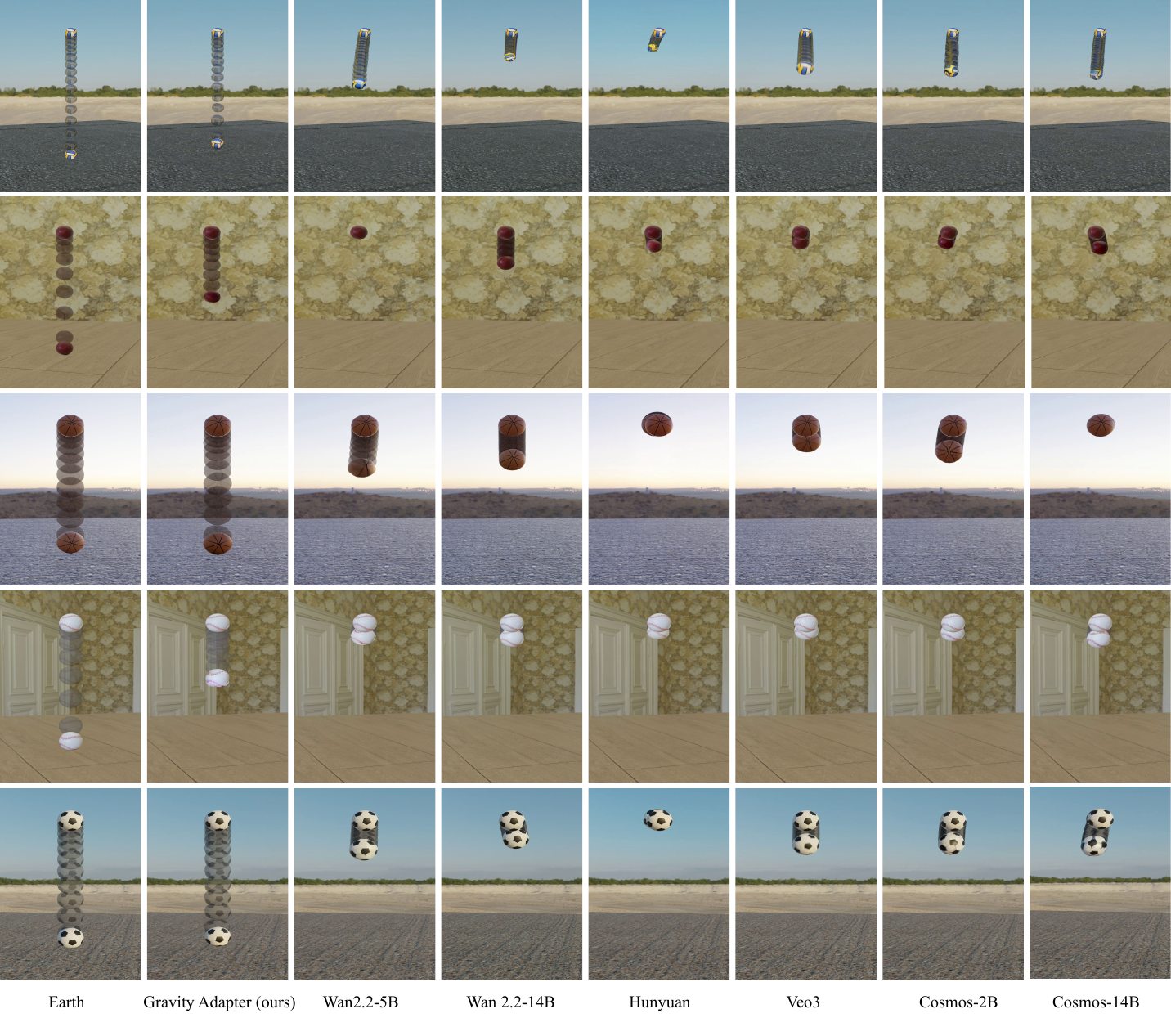}
  \vspace{-10pt}
  \caption{\textbf{Results for Single-ball drops.}
  Stroboscopic composites (left) visualize ball positions at equal time intervals from release. 
  The panels show the trajectories performed by each model during the time it takes a ball falling 
  under $9.81 \, \text{m/s}^2$ to reach the ground, revealing systematic under-acceleration across all models.}
  \label{fig:supple_single_ball_quali}
  \vspace{-10pt}
\end{figure*}

\begin{figure*}[t]
  \centering
  \includegraphics[width=\textwidth,keepaspectratio]{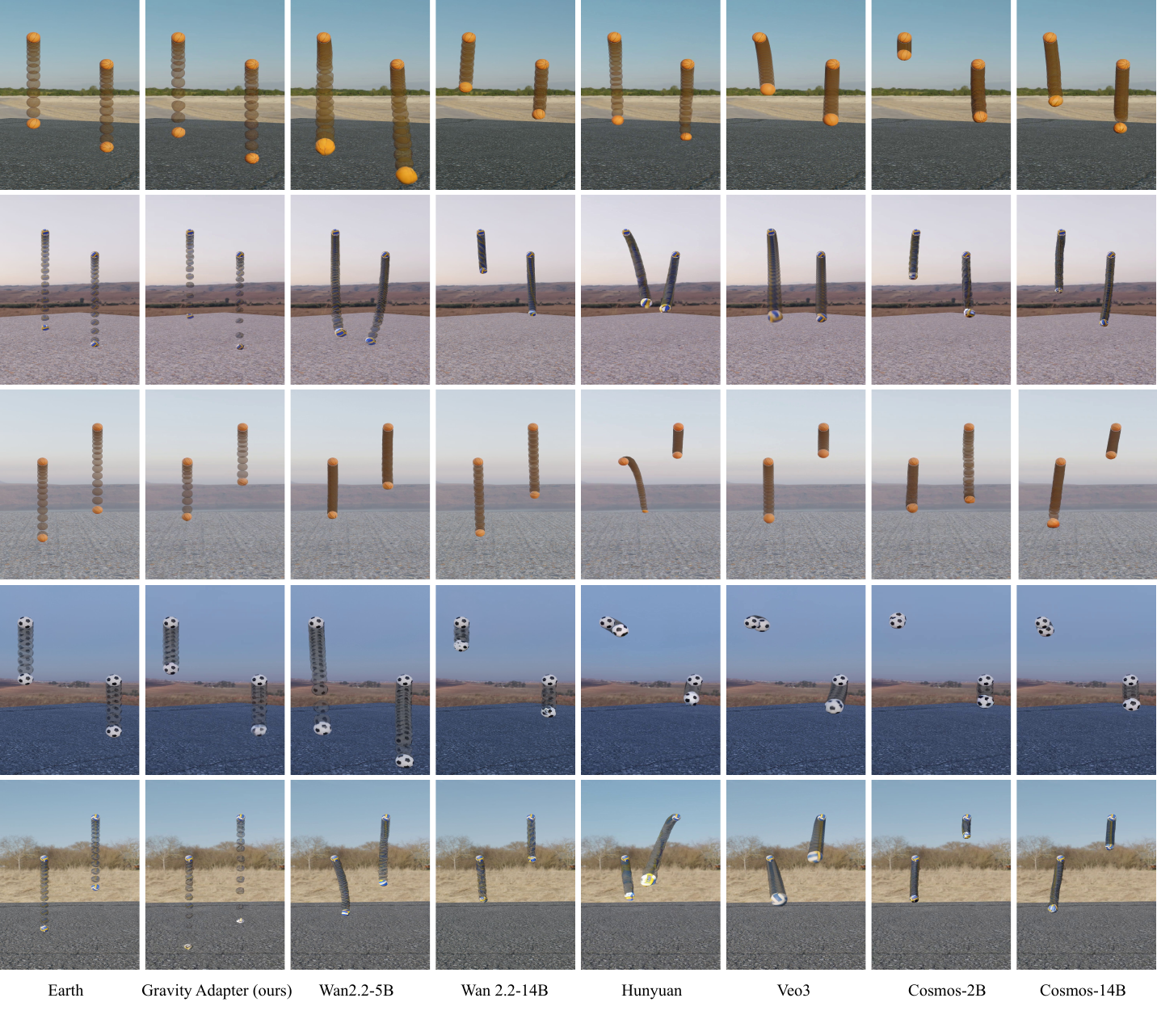}
  \vspace{-10pt}
  \caption{\textbf{Additional Results for Zero-shot generalization to double ball dropping.} }
  \label{fig:supple_double_ball_quali}
  \vspace{-10pt}
\end{figure*}

\begin{figure*}[p]
  \centering
  \includegraphics[width=0.7\textwidth,keepaspectratio]{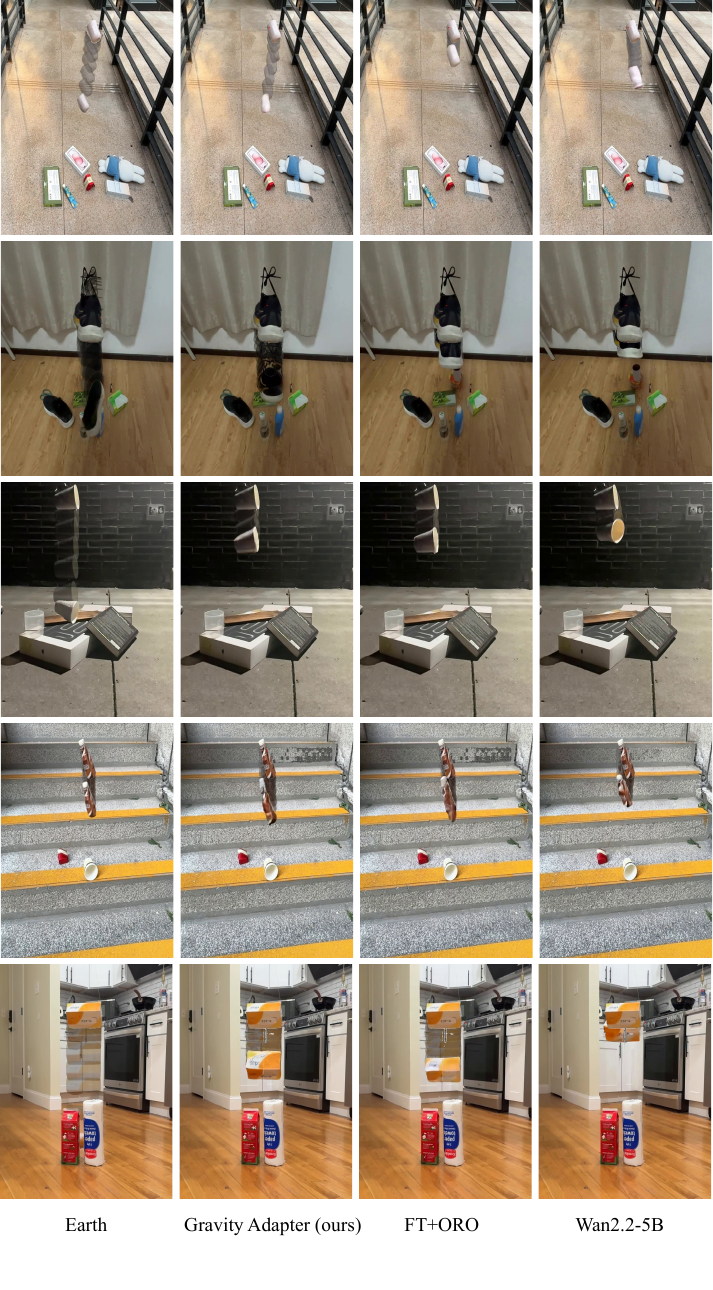}
  \vspace{-10pt}
  \caption{\textbf{Additional Results for Zero-shot generalization to real world scenes~\cite{li2025pisa}.}
  }
  \label{fig:supple_pisa_fig}
  \vspace{-10pt}
\end{figure*}

\begin{figure*}[t]
  \centering
  \includegraphics[width=0.7\textwidth,keepaspectratio]{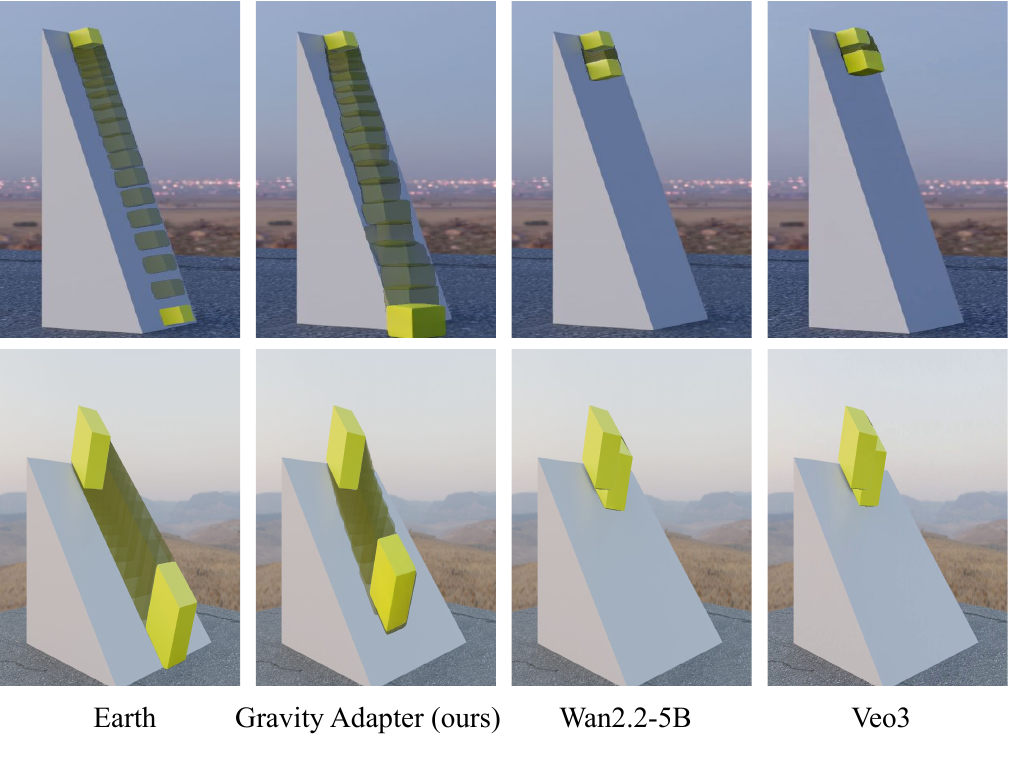}
  \vspace{-10pt}
  \caption{\textbf{Additional Results for Zero-shot generalization to inclined planes.}
   The gravity adapter generalizes to inclined surfaces kept at angles between 30 and 75 degrees.}
  \label{fig:supple_incline}
  \vspace{-10pt}
\end{figure*}

\end{document}